%% file: paper.tex
\documentclass[lettersize,journal]{IEEEtran}
\usepackage{xpatch}
\usepackage{amsmath,amsfonts}
\usepackage{algorithmic}
\usepackage{array}
\usepackage[caption=false,font=normalsize,labelfont=sf,textfont=sf]{subfig}
\usepackage{etoolbox}
\usepackage{textcomp}
\usepackage{stfloats}
\usepackage{url}
\usepackage{verbatim}
\usepackage{graphicx}
\usepackage{multirow}
\usepackage{booktabs}
\usepackage[draft]{todonotes}
\usepackage[inkscapearea=page]{svg}
\usepackage{hyperref}
\usepackage[noabbrev,capitalize]{cleveref}
\usepackage{pifont}
\usepackage[export]{adjustbox}
\usepackage{rotating}
\usepackage{soul}
\usepackage[normalem]{ulem}
\usepackage{ifthen}
\makeatletter
\AtBeginDocument{\patchcmd{\sf@subfloat}%
  {\maincaptiontopfalse}%
  {\maincaptiontoptrue}%
  {}{}}%
\makeatother

\newcommand{\diff}{0}

\ifthenelse{\equal{\diff}{1}}
{
\newcommand{\revone}[1]{{\color{red}{#1}}}
\newcommand{\revtwo}[1]{{\color{blue}{#1}}}
\newcommand{\revgeneral}[1]{{\color{purple}{#1}}}
}
{
\newcommand{\revone}[1]{#1}
\newcommand{\revtwo}[1]{#1}
\newcommand{\revgeneral}[1]{#1}
}

\def\BibTeX{{\rm B\kern-.05em{\sc i\kern-.025em b}\kern-.08em
    T\kern-.1667em\lower.7ex\hbox{E}\kern-.125emX}}
\usepackage{balance}
\begin{document}
\title{Magnifier: A Multi-grained Neural Network-based Architecture for Burned Area Delineation}
\author{Daniele Rege Cambrin, Luca Colomba, Paolo Garza \thanks{Daniele Rege Cambrin, Luca Colomba, Paolo Garza are with Politecnico di Torino, Turin, Italy. (Corresponding author: Daniele Rege Cambrin)}}

\maketitle

\begin{abstract}

In crisis management and remote sensing, image segmentation plays a crucial role, enabling tasks like disaster response and emergency planning by analyzing visual data. Neural networks are able to analyze satellite acquisitions and determine which areas were affected by a catastrophic event. The problem in their development in this context is the data scarcity and the lack of extensive benchmark datasets, limiting the capabilities of training large neural network models. In this paper, we propose a novel methodology, namely Magnifier, to improve segmentation performance with limited data availability. The Magnifier methodology is applicable to any existing encoder-decoder architecture, \revtwo{as it extends a model by merging information at different contextual levels through a dual-encoder approach: a local and global encoder. Magnifier analyzes the input data twice using the dual-encoder approach. In particular, the local and global encoders extract information from the same input at different granularities. This allows Magnifier to extract more information than the other approaches given the same set of input images.} Magnifier improves the quality of the results of +2.65\% on average IoU while leading to a restrained increase in terms of the number of trainable parameters compared to the original model. We evaluated our proposed approach with state-of-the-art burned area segmentation models, demonstrating, on average, comparable or better performances in less than half of the GFLOPs.
\end{abstract}


\section{Introduction}
\label{sec:introduction}

\begin{figure}[htb]
    \centering
    \revone{
        \includegraphics[width=\linewidth]{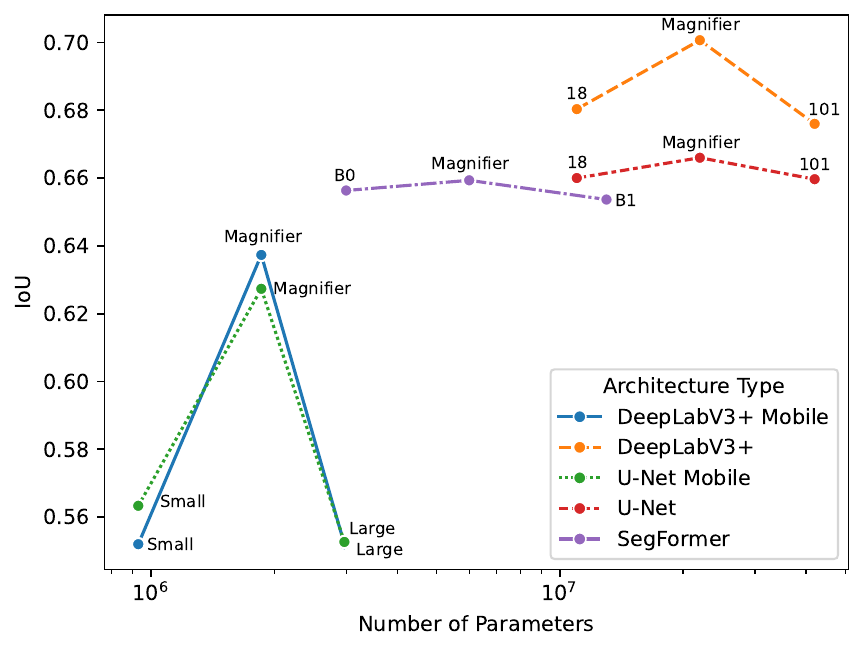}
        \caption{\textbf{Average Mean IoU vs. Number of parameters}. The Average Mean IoU has been computed considering all datasets. \textit{Architecture Type} refers to the family of base networks employed for segmentation. On average, the \textit{Magnifier} backbone achieves IoU improvements compared to MobileNetV3 Small and Large, ResNet-18 and 101, and MiT-B0 and B1 without increasing the number of parameters too much. }
        \label{fig:iou_vs_size}
    }
\end{figure}

Forests represent a fundamental resource for the environment, society, and the entire global ecosystem.
In recent years, an increase in the number of forest wildfires has been observed due to several different causes, such as human activities, climate change, and extreme weather conditions. It is estimated to grow in the next decades~\cite{dupuy2020climate,ruffault2020increased,halofsky2020changing}, endangering the ecosystem and leading to environmental damage and economic and humanitarian losses. Since the restoration process of damaged areas takes several years, losses are observed over the entire time the ecosystem requires to regenerate, leading to collateral damages such as a higher risk of landslides.
Identifying the afflicted areas correctly and planning the post-wildfire restoration process are relevant activities for natural hazard management.
Such a process can be supported by data availability through high-resolution sensors mounted on aircrafts and satellites, which can acquire information at a continental scale. This, adopted in conjunction with the most recent development of machine learning and deep learning techniques, makes the Earth Observation (EO) field particularly interesting to researchers and domain experts to develop automatic monitoring techniques via remote sensing. Such methodologies support local authorities and could improve the handling of catastrophic events and natural hazards. \revtwo{Deep learning models have revolutionized computer vision and image analysis in different application domains, including remote sensing. A few examples of such applications in the remote sensing domain are wildfire severity estimation~\cite{gibson2023image,double_step} and delineation~\cite{california_no_dataset,rafik2021,burned_area_delineation,farasin2020supervised}, flooding detection~\cite{palomba2020sentinel}, air quality estimation~\cite{arnaudo_airquality}, land cover~\cite{landcover_dl, landcover_dl2,zhao2022cnn} and scene classification \cite{Datla2021,Datla2024}.}

In this paper, we tackle the problem of burned area delineation (i.e., identifying areas previously affected by forest wildfires) using data collected by the Sentinel-2 \revone{and Landsat-8 missions'} satellites and deep learning models. Such data and models enable the possibility to provide first post-wildfire damage assessments quickly, compared to the time required by domain experts to identify damaged regions manually or semi-automatically.

The previously proposed deep learning-based approaches for burned area delineation (e.g., \cite{california_no_dataset,rafik2021, burned_area_delineation,farasin2020supervised}) apply general-purpose semantic segmentation models to solve this task. 
Different SOTA deep learning architectures have been used, including CNNs (e.g., U-Net~\cite{unet} or DeepLabV3+~\cite{deeplabv3plus}) 
and vision transformers (e.g., SegFormer~\cite{segformer}).
\Cref{fig:iou_vs_size} reports the average intersection over union (IoU) achieved by such standard semantic segmentation architectures and those of our proposed solution, called Magnifier, applied on \revone{three} open-access datasets.
The results of general-purpose SOTA deep-learning semantic segmentation models are quite satisfactory, particularly when CNN-based architectures are used. However, an uncommon behavior is highlighted in \Cref{fig:iou_vs_size}. Unlike in other semantic segmentation tasks~\cite{deeplabv3plus,segformer,mobilenet}, increasing the model size, considering the same architecture \revone{type}, does not increase or negligibly improves the quality of the results (the results of two models of different sizes for each architecture type are reported in \Cref{fig:iou_vs_size}). This is probably because the available labeled datasets are not sufficiently large to train the larger models, which sometimes start overfitting, impacting the predictions (for example, consider SegFormer B1 vs. SegFormer B0 in \Cref{fig:iou_vs_size}).
Hence, further enhancing the results can be achieved only by increasing the amount of labeled data, which is costly and frequently impracticable, or proposing a different approach to better use the already available labeled data.
Also, we would like to point out that the largest models cannot be used in some scenarios due to resource constraints (e.g., on embedded devices). Hence, new architectures able to increase the quality of the results by limiting the size of the models \revone{without requiring more labeled data} are helpful.
To address such issues, we propose a multi-grained neural network architecture, namely Magnifier, which combines information and features computed at different levels of detail (contextual levels) to perform better than state-of-the-art architectures (i)~using the {\it same amount of labeled data} and (ii)~{\it smaller neural networks}.

Figure~\ref{fig:iou_vs_size} shows how Magnifier, which can be applied on top of well-known deep learning networks, achieves better intersection over union (IoU) than SOTA architectures without the need to use significantly larger models in terms of parameters.

The code of Magnifier is publicly available at \url{https://github.com/DarthReca/magnifier-california}.

Our contributions can be summarized as follows:
\begin{itemize}
    \item We propose a new technique to increase segmentation performance without collecting more labeled data. %
    \item We propose a versatile technique/architecture that can leverage many different encoder-decoder deep-learning models without much effort.
    \item We compared the effectiveness of the proposed technique with previous well-known SOTA convolutional and transformer-based architectures for semantic segmentation.
\end{itemize}

\section{Related Works}
\label{sec:related_works}

This section covers the works related to deep learning, semantic segmentation, and their application in earth observation and burned area delineation.

\subsection{Deep Learning}
Deep learning is a subset of machine learning that uses artificial neural networks with multiple layers to model and understand complex patterns in data \cite{deep_learning}.
The primary innovation of neural networks and deep learning is the ability to automatically learn hierarchical feature representations from raw data, eliminating the need for manual feature engineering. These models, especially Convolutional Neural Networks (CNNs) \cite{cnn}, Recurrent Neural Networks (RNNs) \cite{lstm,brnn}, and Transformer \cite{transformer} architectures, can process vast amounts of data and extract meaningful patterns at multiple levels of abstraction. CNNs, for example, are particularly well-suited for image processing tasks, while RNNs and Transformers are suited for sequential data like time series and language. More recently, the latter architecture demonstrated its superior performances also in the computer vision domain~\cite{vit} in the presence of a large amount of training data. The hierarchical feature representation learning performed by neural networks is formulated as an optimization problem: given the predicted outcomes by the model and the ground truth information, a loss function is evaluated and minimized throughout the training process \cite{deep_learning}. 
Deep learning has gained widespread attention because it successfully outperforms traditional machine learning algorithms in complex tasks. This success is largely attributed to the availability of large datasets, advances in computational power (particularly GPUs), and innovations in neural network architectures that enable these models to handle more diverse and intricate data types. It has become a foundational technology for researchers in different fields, such as computer vision \cite{mobilenet}, natural language processing \cite{bert}, and reinforcement learning \cite{dqn}.

\subsection{Semantic Segmentation}
Semantic segmentation is a fundamental task in computer vision and image processing that involves classifying each pixel in an image based on its semantic meaning.

Classical methods in semantic segmentation have laid the foundation for developing more sophisticated techniques. For example, the usage of histogram analysis\cite{otsu1979threshold}, graph partition techniques \cite{normalized_cuts}, and mean shift\cite{mean_shift} provided promising results in segmenting with a certain number of limitations until the advent of deep learning.

The employment of Convolutional Neural Networks (CNNs) revolutionized the computer vision field by solving various tasks~\cite{tof} and providing outstanding results in dealing with complex segmentation problems. Networks such as the multiple versions of DeepLab~\cite{deeplab,deeplabv3} and UNet~\cite{unet} have proven to be quite capable in many different fields, ranging from medical to satellite imagery\cite{ColombaCikm22}. 

The last advancements in computer vision propose new vision transformer architectures such as Swin Transformer~\cite{swin} and SegFormer\cite{segformer}, which provide new state-of-the-art results in semantic segmentation benchmarks.

While well-known encoders like ResNets\cite{resnet} and SegFormer\cite{segformer} create representations at different resolution levels, they do not exploit different contextual levels. Some other works acted directly onto the input image resolution without changing also, in this case, the granularity of the context~\cite{fcn_multiple_res}. As described in the following paragraphs, our work distinguishes itself against the latter due to the presence of multiple encoders to handle different information granularity, whereas the aforementioned paper adopts a Siamese U-Net model to analyze the original input image and a downsampled version (i.e., duplicated information is seen four times). 

The Magnifier architecture we propose in the paper is built on top of these established neural networks. It effectively combines models applied to images at different ``granularities'' to increase their performances without increasing the number of labeled input images,
which would require time and resources. Since satellite imagery covers vast areas and contains coarse and fine details, combining global and local views leads to more informative embeddings without requiring more labeled data and larger models.

\subsection{Earth Observation}
Recently, researchers started investigating the applicability of state-of-the-art machine learning models to hyperspectral data (e.g., satellite data) in the remote sensing domain. In fact, considering the field of Earth Observation, several tasks can be formulated as common machine learning problems, such as land cover classification~\cite{bigearthnet,wang2019scene}, crop classification~\cite{winter_wheat_mapping,faqe2023improving,cambrin2024},\revtwo{image segmentation~\cite{demir2018deepglobe,Swetha2023,Datla2022}}, and visual question answering~\cite{vqa_dataset}. Over the years, the amount of labeled data has risen, enabling the possibility of training better and larger machine learning models. This latter point, combined with a greater availability of computational resources, led to the improvement of deep learning methodologies in many different fields, including computer vision and hyperspectral image analysis. 
Despite this, focusing on the context of resource and emergency management as well as disaster recovery, the data availability problem still persists due to the rarity of such events. As a consequence, this problem requires researchers to develop well-performing machine learning methodologies with a limited amount of data. Even with such a limit, machine learning methodologies demonstrated superior performances compared to traditional techniques. In~\cite{palomba2020sentinel}, tree-based and UNet algorithms surpassed index-based analyses of flooded areas. Similarly,~\cite{mmflood} tested several different architectures in flood segmentation and compared them against Otsu, whereas~\cite{double_step} evaluated the aforementioned neural network architecture against the dNBR index in the burned area delineation problem.

\subsection{Burned area delineation}
\subsubsection{Burned area delineation employing spectral indices}
The burned area delineation problem is a well-known problem in the field of EO. 
Several works tackle the problem by analyzing the spectral signature of the affected areas (e.g.,~\cite{spectral_signature_analysis,stroppiana2012method,martin2006burnt}). Domain experts developed spectral indices, also known as burned area indices, which are sensitive to vegetation and humidity to distinguish damaged and undamaged areas given a satellite acquisition. Such indices are computed by different combinations of spectral bands, obtaining a single-channel image that can highlight areas affected by the catastrophic event under consideration. This is possible because different bandwidths collect information strictly related to vegetation, making identifying the areas of interest feasible. Examples of such indices are BAIS2~\cite{bais2}, Normalized Burn Ratio~\cite{nbr_cit}, and Normalized Burn Ratio 2~\cite{nbr2_cit}. The resulting index is consequently analyzed for solving image segmentation by applying threshold-based techniques. One of the main issues of the proposed approaches is that, given a burned area index, identifying a unique threshold valid worldwide, which adequately segments burned areas, is difficult due to different lighting conditions, vegetation, terrain characteristics, and morphological features~\cite{not_unique_thresholds}. Identifying the threshold for a single region is also a non-trivial task. For this reason, automatic deep-learning models are more effective and preferred.

\subsubsection{Burned area delineation employing deep learning models}
More recently, with the development of modern computer vision techniques based on the adoption of deep learning methodologies, researchers started the investigation and the application of neural networks also in the field of EO, demonstrating state-of-the-art performances~\cite{audebert2017semantic, LANG2022112760}. Several deep models are currently adopted, demonstrating their effectiveness in analyzing satellite imagery and processing multispectral acquisitions. More specifically, different researchers successfully applied convolutional neural networks~\cite{brand2021semantic, rs13081509} and transformer-based architectures~\cite{HU2023228} to the burned area delineation problem. From the latter work, UNet-based models demonstrated state-of-the-art performances in the burned area identification domain~\cite{monaco21}. Similarly,~\cite{burntnet} extended the U-Net architecture by enhancing the convolutional blocks in both the encoder and decoder part of the network with multi-patch multi-level residual morphological blocks (MP-MRM), preserving the skip connections typical of such segmentation model. Internally, the MP-MRM blocks leverage skip-connections, erosion, and dilation operators with a quadratic structuring function, as well as traditional convolutional layers, to solve the burned area delineation problem.
The architectures designed so far rely on a significant amount of data (collected from different sources, in some cases~\cite{california_no_dataset}) and use large general-purpose semantic segmentation models.
Solutions based on many labeled data, multiple sources, or computationally expensive models could be challenging to scale to real applications.
We propose a multi-grained neural network architecture to reduce complexity and computational costs and improve the quality of the results 
without requiring more labeled data and using ``small-size'' models. Using the proposed architecture shows how smaller models can be competitive against more complex ones using the same amount of input data. Good performances combined with the necessity of fewer resources can boost the use of deep learning-based architectures in the EO domain.

\section{Materials and Method}
\label{sec:methodology}
In this section, we formally present the problem of burned area delineation, the dataset used for the study, the employed architecture, and finally, the loss function for the neural network training and the metrics.

\subsection{Problem Statement}
\label{sec:prob_stat}

This work addresses the burned area segmentation problem based on post-fire images. The task can be formalized as follows.

Let $I$ be an arbitrary satellite image of size $W \times H \times D$, where $W$ and $H$ are the width and height of the images in pixels, respectively, while $D$ is the depth of the images (i.e., the number of features per pixel). 
The objective is to automatically create a binary mask $M$ of size $W \times H$ associated with $I$, where the value 1 indicates the associated pixel contains a burned area. In contrast, 0 is related to unburned pixels.

To solve the problem, we have at our disposal a set $S_L$ of labeled images (with the same features of $I$) for which the associated masks (with the same characteristics of $M$) are known. The labeled set of images $S_L$ is used to train a semantic segmentation model $SM$.
After the training, $SM$ can be used to predict the mask $M$ of an arbitrary (new) image $I$ for which the associated mask is unknown.

\subsection{Datasets}
\label{sec:dataset}

\revone{
\begin{table}[b]
    \caption{Characteristics of the used datasets.}
    \label{tab:dataset_comparison}
    \resizebox{\linewidth}{!}
    {
        \begin{tabular}{l|lll}
        \toprule
         & CaBuAr \cite{cabuar} & Europe \cite{double_step} & Indonesia \cite{prabowo2022} \\
        \midrule
        Resolution & 20m & 10m & 30m \\
        Channels & 12 & 12 & 8 \\
        Forest Fires & 340 & 73 & 81 \\
        Start-End date & Jan, 2015 - Dec, 2022 & July, 2017 - July, 2019 & Jan, 2019 - Dec, 2021 \\
        Burned surface & $\sim 11000$ km$^2$ &  $\sim 2000$ km$^2$ & $\sim$7000 km$^2$\\
        Number of images & 688 & 449 & 227 \\ 
        \bottomrule
        \end{tabular}
    }
\end{table}
}

We evaluated our methodology on \revone{three} publicly available datasets spanning three different areas of the globe: California\footnote{Available at \url{https://huggingface.co/datasets/DarthReca/california_burned_areas}}~\cite{cabuar}, Europe\footnote{Available at \url{https://zenodo.org/record/6597139}}~\cite{double_step} \revone{and Indonesia\footnote{Available at \url{https://data.mendeley.com/datasets/fs7mtkg2wk}} \cite{prabowo2022}}. \revone{See \Cref{tab:dataset_comparison} for an overview of their characteristics.}

Both California and Europe datasets use the Sentinel-2 mission of the European Space Agency (ESA). It comprises a pair of identical satellites orbiting Sun-synchronously at an average height of 786 km. The mission was designed to contribute to Land Monitoring, Emergency Response, and Security services. Each satellite is equipped with the Multi-Spectral Instrument (MSI) with 13 spectral bands spanning from the visible and the near-infrared to the short wave infrared with resolution from 10m to 60m, as shown in \Cref{tab:spectral_bands}. Many of them are sensitive to vegetation, permitting the monitoring of green areas of the globe. The field of view is 290 km. They provide a mean global revisiting time smaller than five days \cite{sentinel2}. 

ESA makes available to final users L1C and L2A products. These products comprise ortho-images in UTM/WGS84 projection, each covering a surface area of $110 \times 110$ km$^2$. L2A products provide atmospherically corrected Surface Reflectance (SR) images derived from the associated Level-1C products. The atmospheric correction process involves several essential steps to account for various atmospheric effects. These steps include correcting for the scattering of air molecules, known as Rayleigh scattering, and compensating for the absorbing and scattering effects caused by atmospheric gases (e.g., ozone, oxygen, and water vapor) and aerosol particles \cite{l2a}. This process results in the omission of band 10 from the final product, while L1C comprises all available spectral bands.

\begin{table}[htb]
    \centering
    \caption{Sentinel-2 spectral bands}
    \label{tab:spectral_bands}
    \resizebox{\linewidth}{!}{\input{tables/s2-bands}}
\end{table}

\revone{
The Indonesia dataset uses the Landsat-8 mission of the National Aeronautics and Space Administration (NASA). It is comprised of a satellite equipped with the Operation Land Imager (OLI) and the Thermal Infrared Sensor (TIRS). OLI captures nine visible and infrared spectral bands with a resolution of 30m, while TIRS has two thermal infrared bands at 100m resolution as detailed in \Cref{tab:l8_spectral_bands}. The revisit time is 16 days. It aims to develop a scientific understanding of the Earth system  \cite{roy2014}. NASA makes L1GT, L1T, L1TP, and L2SR products available to final users. These products comprise ortho-images in UTM/WGS84 projection, each covering a surface area of $185 \times 185$ km$^2$. L2SR products provide atmospherically corrected Surface Reflectance (SR) images derived from the associated L1T or L1TP products.
Hence, this dataset is composed of satellite images/input products with bands and characteristics different from those of the other two datasets. These differences do not impact the applicability and quality of our methodology.

\begin{table}[htb]
    \centering
    \revgeneral{
        \caption{Landsat-8 spectral bands}
        \label{tab:l8_spectral_bands}
        \resizebox{\linewidth}{!}
        {
            \begin{tabular}{l|c|c|l}
                \toprule
                Band & Resolution & Central wavelength & Description \\ \midrule
                B1 & 30 m & 440 nm & Ultra Blue (Coastal and Aerosol)\\
                B2 & 30 m & 480 nm & Blue \\
                B3 & 30 m & 560 nm & Green \\
                B4 & 30 m & 650 nm & Red \\
                B5 & 30 m & 860 nm & Near Infrared (NIR) \\
                B6 & 30 m & 1610 nm & Short Wave Infrared (SWIR)\\
                B7 & 30 m & 2200 nm & Short Wave Infrared (SWIR)\\
                B8 & 15 m & 590 nm & Panchromatic \\
                B9 & 30 m & 1370 nm & Cirrus \\ 
                B10 & 100 m & 10900 nm & Thermal Infrared \\
                B11 & 100 m & 12000 nm & Thermal Infrared \\\bottomrule
            \end{tabular}
        }
    }
\end{table}
}

\subsubsection{CaBuAr Dataset}
The CaBuAr (California Burned Areas) dataset \cite{cabuar} comprises L2A products of Sentinel-2 and ground truth masks provided by the California Department of Forestry and Fire Protection. In \Cref{fig:firemaps}a, it is possible to look at the geographical distribution of the analyzed wildfires over California. The dataset is composed of 340 wildfires that happened between January 2015 and December 2022. The total burned area surface is more than 11000 km$^2$. The post-fire acquisition is downloaded with at most a month delay after the wildfire containment date. If necessary, the bands were upsampled or downsampled to 20m resolution to make the matrix tractable by a neural network. The dataset comprises 688 images of size $512 \times 512 \times 12$ divided into five folds. A complete overview of the characteristics can be seen in \Cref{tab:dataset_comparison} and a sample can be seen in \Cref{fig:datasets_samples}a.

\subsubsection{Europe Dataset}
\begin{figure*}[htb]
    \revone{
        \centering
        \subfloat[California]{\includegraphics[width=0.3\linewidth]{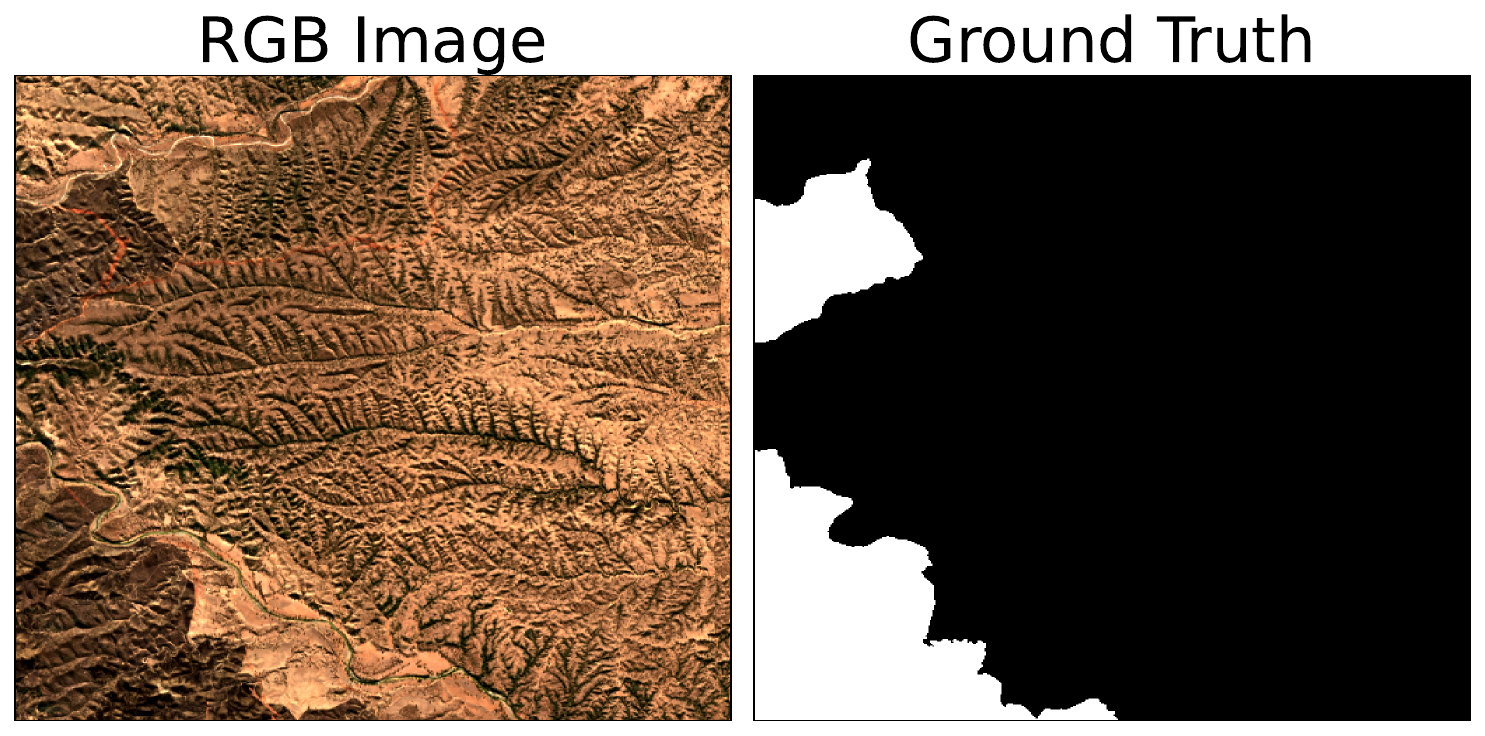}}
        \hfil
        \subfloat[Europe]{\includegraphics[width=0.3\linewidth]{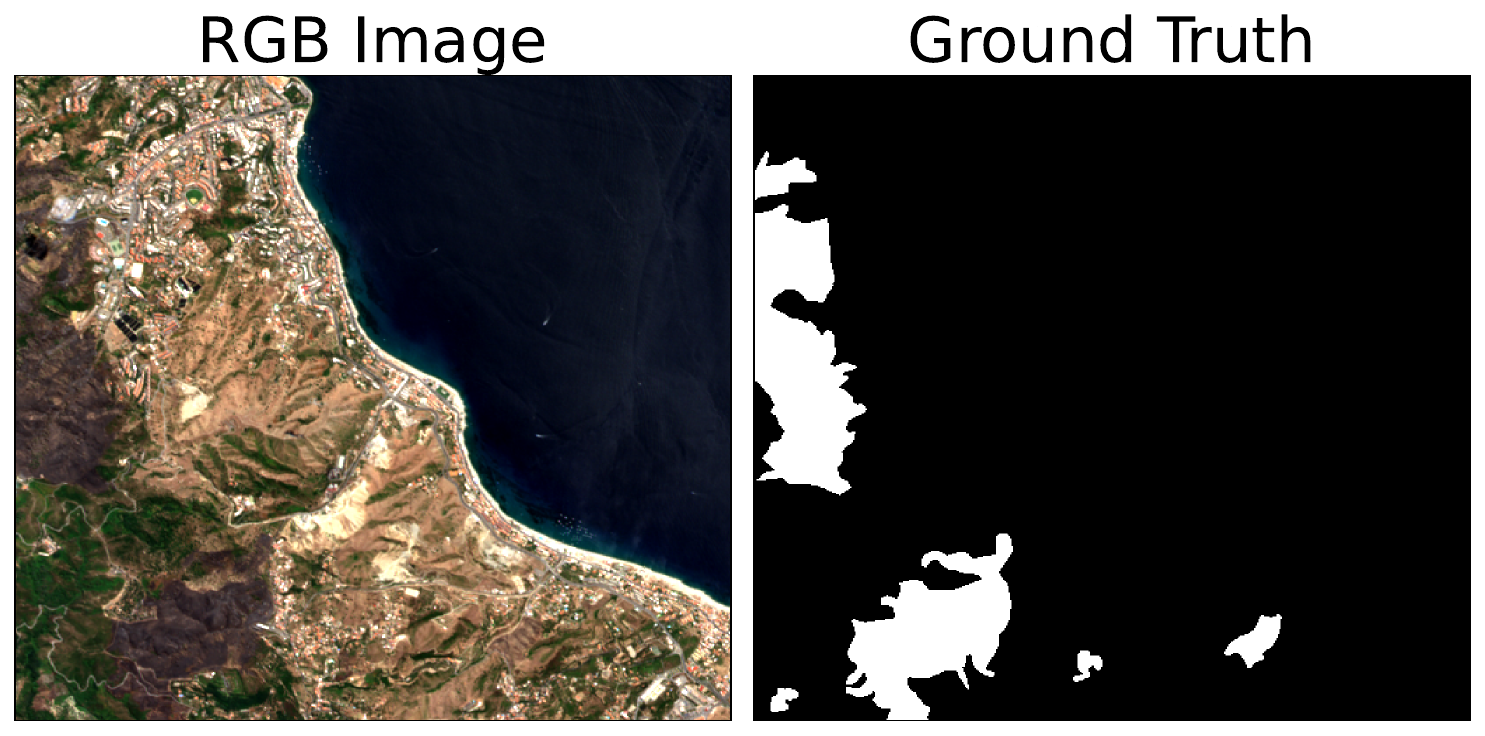}}
        \hfil
        \subfloat[Indonesia]{\includegraphics[width=0.3\linewidth]{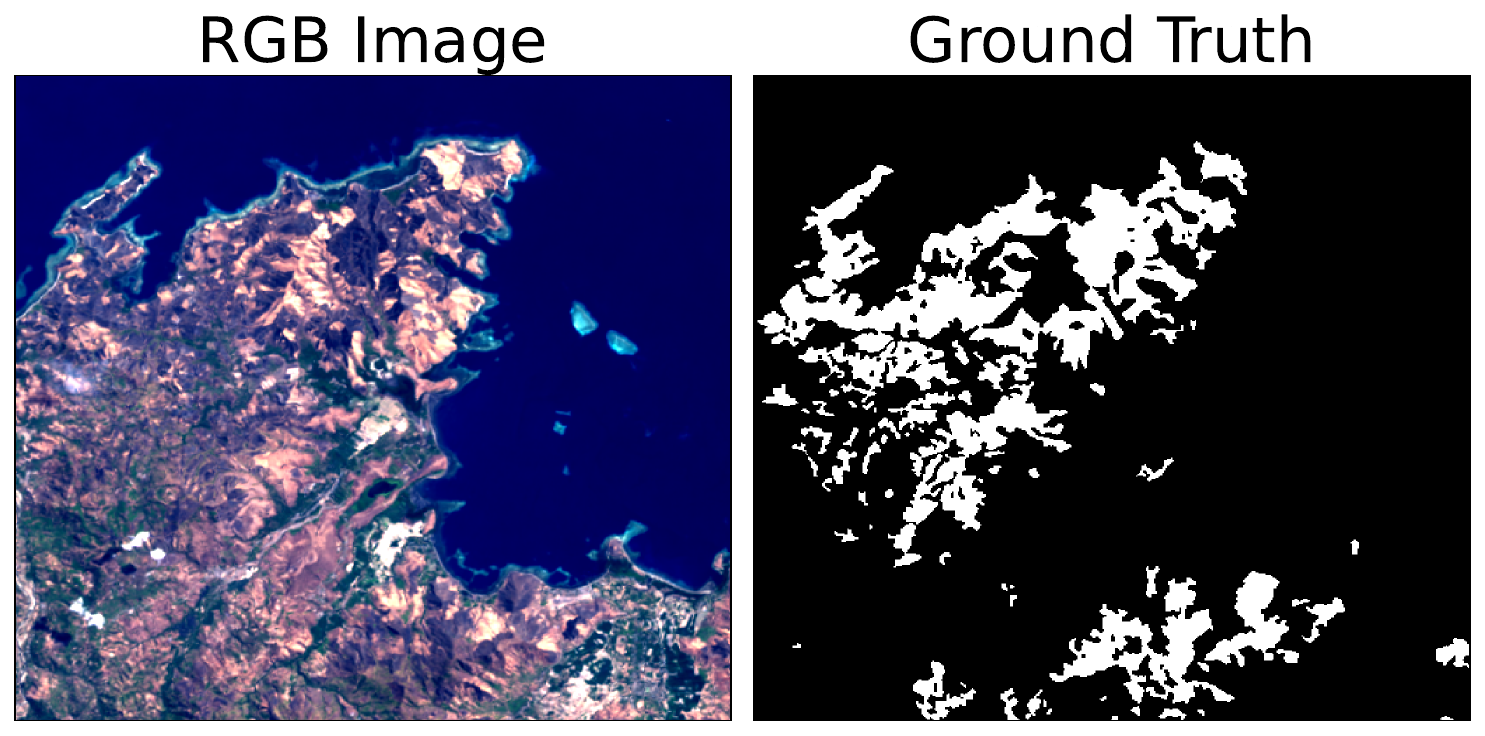}}
        \caption{RGB samples taken from the three datasets with the corresponding binary ground truth.}
        \label{fig:datasets_samples}
    }
\end{figure*}

\begin{figure*}
    \centering
    \subfloat[California \cite{cabuar}]{\frame{\includegraphics[height=4cm]{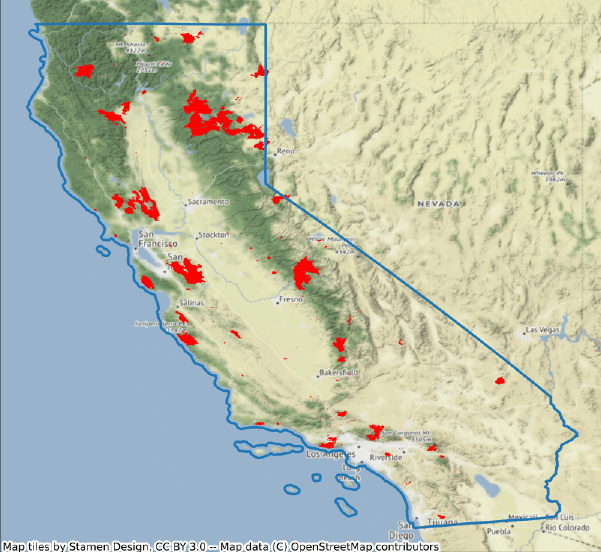}}}
    \hfil
    \subfloat[Europe \cite{double_step}]{\frame{\includegraphics[height=4cm]{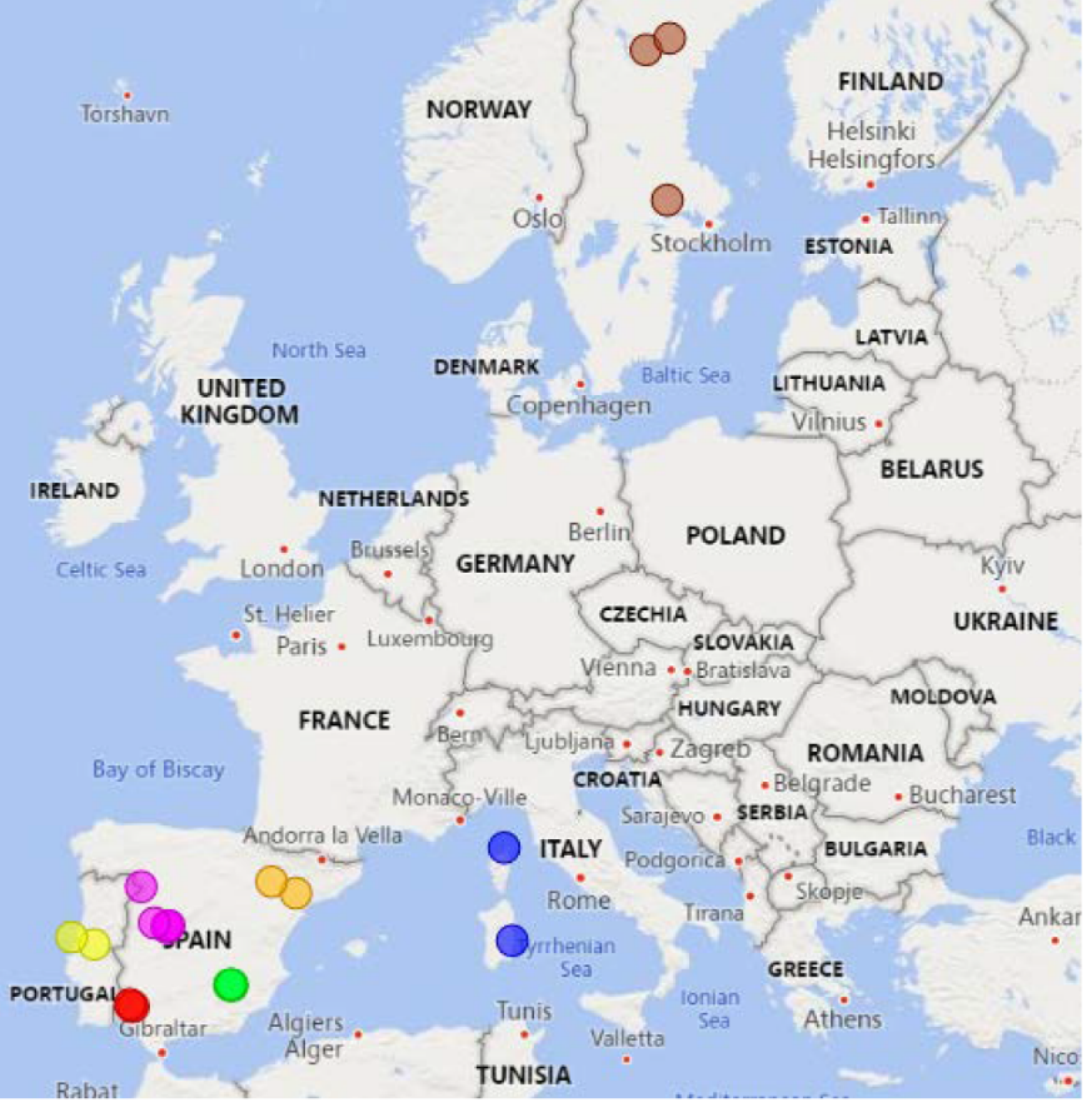}}}
    \hfil
    \revone{
    \subfloat[Indonesia \cite{prabowo2022}]{\frame{\includegraphics[height=4cm,width=0.5\linewidth]{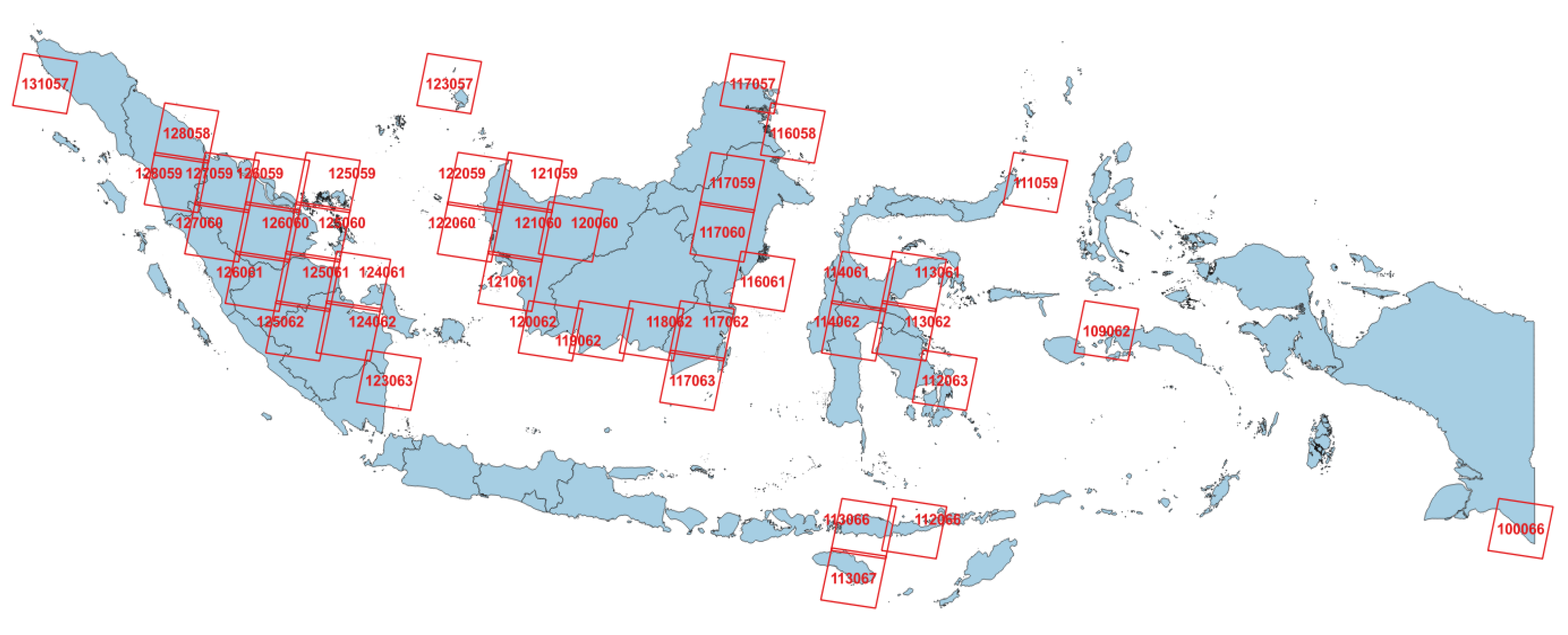}}}
    }
    \caption{Distribution of wildfires in the analyzed datasets. In (a) the areas covered by wildfires in \textit{red}. In (b), the location of the wildfires, where each color represents the fold it belongs to. In (c), the areas covered by the dataset are in red squares.}
    \label{fig:firemaps}
\end{figure*}

The Europe dataset \cite{double_step} comprises L2A products of Sentinel-2 and Sentinel-1 and ground truth provided by the Copernicus Emergency Management System. In \Cref{fig:firemaps}b, it is possible to look at the geographical distribution of the considered wildfires. The dataset is composed of 73 wildfires that happened between July 2017 and July 2019. The total burned area is more than 2000 km$^2$. The timeframe between the containment date and data acquisition is not specified, although the delineation maps were acquired after the wildfire event. All the bands are upsampled to 10m resolution when necessary. The dataset comprises 449 images of size $512 \times 512 \times 12$, divided into seven folds. Its main characteristics are reported in \Cref{tab:dataset_comparison} and a sample can be seen in \Cref{fig:datasets_samples}b.

\revone{
\subsubsection{Indonesia Dataset}
The Indonesia dataset \cite{prabowo2022} consists of 227 manually annotated satellite images derived from Landsat-8 of size $512 \times 512 \times 8$ pixels. Four experts annotated the data, and three other experts evaluated the annotations. In \Cref{fig:firemaps}c, we show the area covered by the dataset.  The dataset is composed of wildfires between 2019 and 2021. The total burned area is more than 7000 km$^2$. The timeframe between the containment date and data acquisition is not specified, although the delineation maps were acquired after the wildfire event. A complete characterization is reported in \Cref{tab:dataset_comparison} and a sample can be seen in \Cref{fig:datasets_samples}c.

The three datasets cover different areas and allow us to analyze the general applicability of the proposed architecture in various contexts. Although CaBuAr focuses only one state (California), it covers the largest area ($\sim11000$ km$^2$ of the burned surface), followed by Indonesia ($\sim7000$ km$^2$) and Europe ($\sim2000$ km$^2$) (see \Cref{tab:dataset_comparison}).
Furthermore, California is characterized by various and complex geology, from high mountains to broad valleys.
}

\begin{figure*}
    \centering
    \revtwo{
    \includegraphics[width=\linewidth]{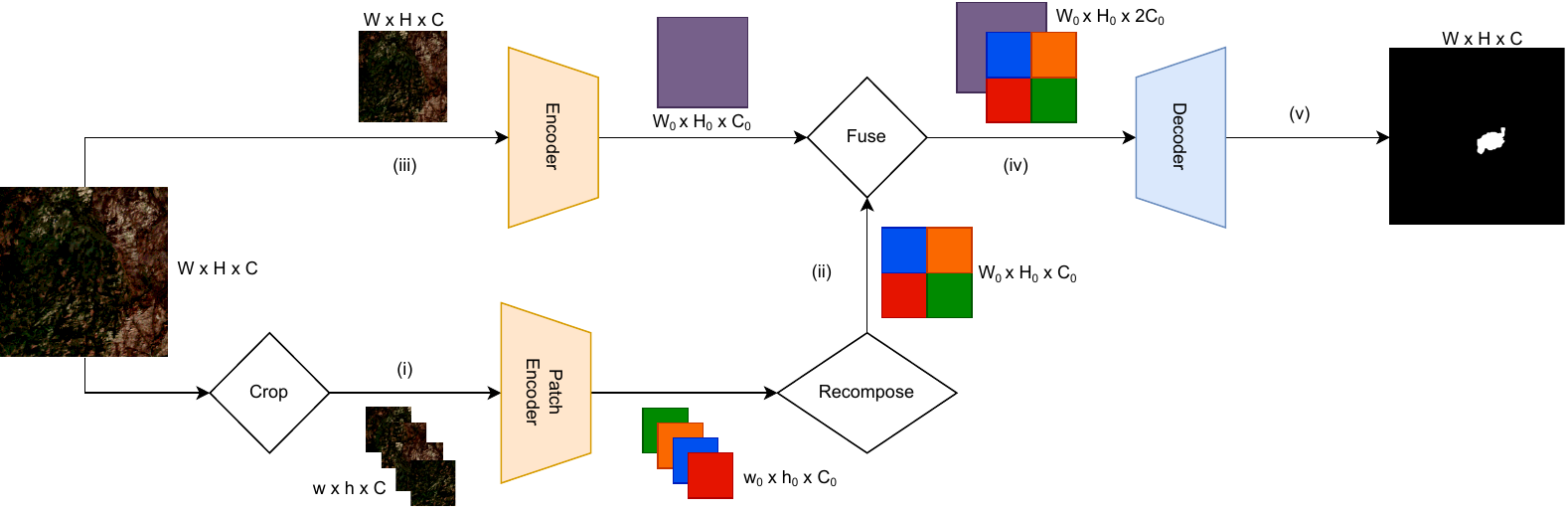}
    }
    \caption{Magnifier architecture. In the lower branch, (i)~the image is cropped in smaller patches \revtwo{(as shown in \Cref{fig:cropping})}, giving each patch to an encoder. (ii)~The encodings are concatenated by putting each one in the original position in the image \revtwo{(as shown in \Cref{fig:recompose})}. In the upper branch (iii), the entire image is given to an encoder. (iv) The two encodings are concatenated along the channel axis, and (v) they are given to the decoder to get the final prediction.}
    \label{fig:magnifier}
\end{figure*}

\subsection{Neural Network Architecture}
The main research question addressed by this paper is whether or not it is beneficial to a neural network-based architecture to fuse information extracted from different patch sizes extracted from $SL$, consequently providing merged contextual information at different granularity to the decoder of an encoder-decoder deep neural network, to perform per-pixel classification (i.e., semantic segmentation). For this reason, we propose an architecture based on two ``paths'' associated with patches of different sizes, extracted from the same image collection, that are fused before being fed to the decoder.
The architecture is quite versatile since it can be applied to various well-known segmentation models in the literature. The main requirement is the model has to be based on an encoder-decoder structure (i.e., encoders and decoders of SOTA encoder-decoder semantic segmentation models can be used in Magnifier).

The Magnifier architecture is presented in \Cref{fig:magnifier}. \revone{Here, we present its general architecture, which is independent of the encoders and decoders used to implement it. The details of the encoders and decoders used/considered to instantiate the Magnifier are reported in the following sections.}

The input of the Magnifier architecture consists of size $W \times H \times C$ images.
The encoder part comprises two branches, each associated with a different encoder and patch size. Finally, a single common decoder is trained to perform the predictions after a fusion step. One encoder receives as input the original (``coarse-grained'') images of size $W \times H \times C$, while the other one (called \textit{Patch Encoder}) accepts (``fine-grained'') images of size $w \times h \times C$, where $w < W$ and $h < H$. We chose non-overlapped patches for simplicity, with $w$ and $h$ sub-multiples of $W$ and $H$, respectively. The two encoders share the same model (e.g., a ResNet encoder) but not the weights. 
The first path is used to learn ``global'' information, while the second learns ``local'' knowledge.

\begin{figure}
    \centering
    \includegraphics[width=\linewidth]{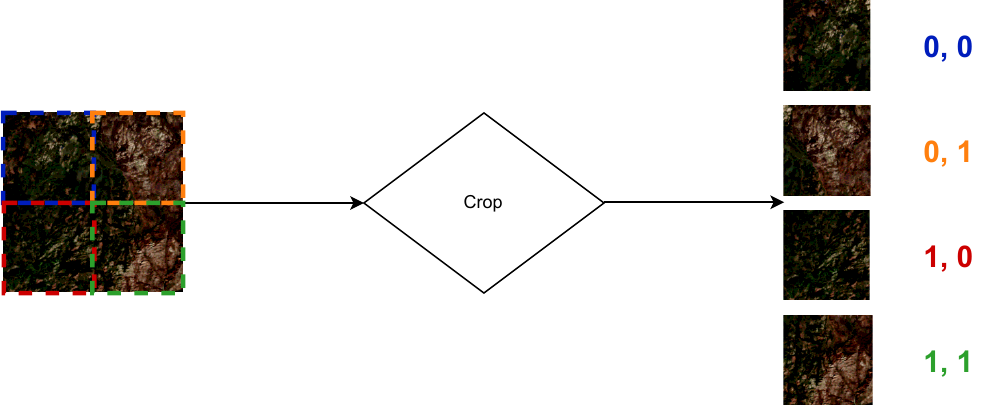}
    \caption{\textbf{Cropping procedure.} The image is cropped in patches, and each of them keeps the original position associated.}
    \label{fig:cropping}
\end{figure}

\begin{figure}
    \centering
    \includegraphics[width=\linewidth]{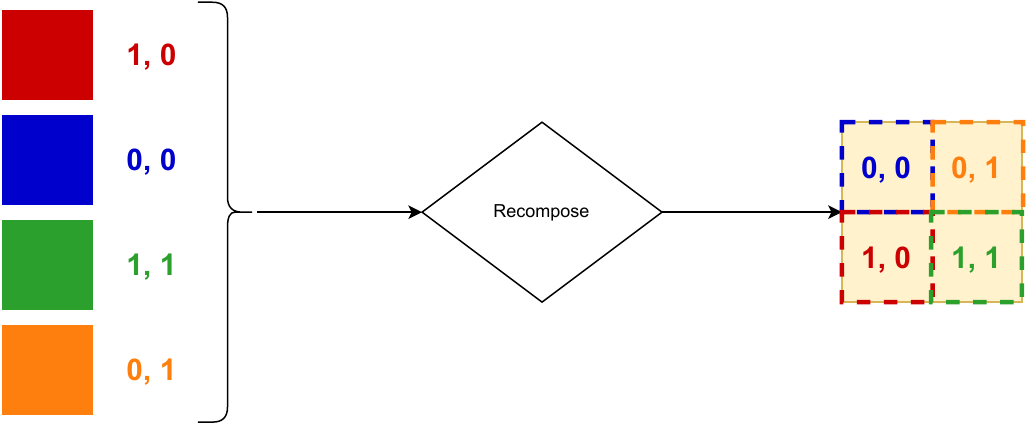}
    \caption{\textbf{Recompose procedure.} The encodings of an image are merged into a single embedding matrix using the position information.}
    \label{fig:recompose}
\end{figure}

The input images are split into smaller patches to feed the second path. Specifically, given an input image $I$ of size $W \times H \times C$, $I$ is split/cropped into several smaller patches $P = \{P_0, P_1 ...\}$ of size $w \times h \times C$. Each patch is coupled with the information about its position in the original image as shown in \Cref{fig:cropping}. For this procedure, the original input image $I$ can be seen as a grid of size $N \times M$ composed of $w \times h \times C$ patches. 
So the information about the position of each patch in the original image $I$ can be stored as a tuple $(\text{ROW}, \text{COLUMN})$, where $\text{ROW} \in \{0, \dots, M-1\}$ and $\text{COLUMN} \in \{0, \dots, N-1\}$. The position $(0, 0)$ was fixed at the top-left corner. $\text{ROW}$ increases from left to right, and $\text{COLUMN}$ from top to bottom. 

The original input image $I$ is embedded into a feature vector of size $W_0 \times H_0 \times C_0$ with the first encoder (top branch in \Cref{fig:magnifier}). Parallelly, each one of the patches $P_i \in P$ is embedded into a feature vector of size $w_0 \times h_0 \times C_0$ with the \textit{Patch Encoder} (bottom branch in \Cref{fig:magnifier}). 

The feature vectors associated with the patches $P_i$ are then concatenated into a global feature vector as shown in \Cref{fig:recompose}, preserving the original order of the patches. For instance, a patch associated with position $(0, 0)$ has its encoding placed in position $(0, 0)$ of the embedding matrix. In this way, we obtain another embedding of size $W_0 \times H_0 \times C_0$ from the second branch of Magnifier.

Consequently, the two feature vectors of sizes $W_0 \times H_0 \times C_0$, extracted by the two encoders, are fused by concatenation along the channel axis ($C_0$).
The resulting embedding is provided as input to the shared decoder to obtain the final prediction.

The Magnifier architecture is general and can be applied on top of different encoder-decoder deep learning models. For instance, the encoder and decoder of a CNN can be used, or those of a vision transformer. 
More details about which SOTA deep learning encoders and decoders have been \revone{used to instantiate the Magnifier and test it} are reported in Section~\ref{sec:basemodels}. 
The source code of Magnifier is publicly available at \url{https://github.com/DarthReca/magnifier-california}.

\subsection{Loss Function}
We employed the Asymmetric Unified Focal (AUF) loss~\cite{yeung_2022}, being the problem highly imbalanced. This loss combines cross-entropy-based and dice-based losses in a single formulation with few hyper-parameters. The asymmetric version of the loss removes the focal suppression to the rare class, giving it more importance. 

The formulation of Asymmetric Unified Focal loss ($L_{AUF}$) is:
\begin{equation}
    L_{AUF} = \lambda L_{maF} + (1 - \lambda) L_{maFT}
\end{equation}
where $\lambda \in [0, 1]$ determine the weights of the two components of the loss, which are modified asymmetric Focal loss ($L_{maF}$) and modified asymmetric Focal Twersky loss ($L_{maFT}$).

The modified asymmetric Focal loss ($L_{maF}$) is defined as:
\begin{equation}
    L_{maF} = -\frac{\delta}{N} y_{i:r}\log p_{t,r} - \frac{1 - \delta}{N} \sum_{c \ne r} (1 - p_{t, c})^{\gamma} \log p_{t,r}
\end{equation}
where $N$ is the number of pixels, $p_{t,r}$ and $p_{t,c}$ are the probability the sample belongs to the rarest and most common classes respectively, $y_{i:r}$ is the ground truth label for the sample belonging to the rare class, $\delta$ is the weighting factor of the components of the loss and $\gamma$ is the suppression factor for the most common class.

The formulation of modified asymmetric Focal Twersky loss ($L_{maFT}$) is:
\begin{equation}
    L_{maFT} = \sum_{c \ne r} (1 - mTI) + \sum_{c = r} (1 - mTI)^{1 - \gamma}
\end{equation}
where $mTI$ is the modified Twersky Index, and $\gamma$ is the enhancement factor applied to the rare class.

The modified Twersky Index is defined as:
\begin{equation}
    mTI = \frac{ \sum_{i}^N p_{r,i} g_{r,i} }{ \sum_{i}^{N} p_{r,i} g_{r,i} + \delta \sum_{i}^{N} p_{r,i} g_{c,i} + (1 - \delta) \sum_{i}^N p_{c,i} g_{r,i}} 
\end{equation}
where $p_{r,i}$ and $p_{c,i}$ are the probability of a pixel belonging to the rarest and most common classes, respectively, and $g_{r,i}$ and $g_{c,i}$ are the ground truth values for the sample belonging to the rare and most common classes respectively.

\section{Experiments}
\label{sec:experiments}

In our experiments, we consider as base model architectures, on top of which Magnifier is applied, two different families of CNN architectures (U-Net~\cite{unet} and DeepLabV3+~\cite{deeplabv3plus}) and a vision transformer (SegFormer \cite{segformer}). 
Using different base model architectures \revtwo{(i.e., using various types of encoders and decoders in Magnifier - \Cref{fig:magnifier})}, we show the effectiveness of the Magnifier approach independently of the models used to instantiate its encoders and decoder.

To evaluate the contribution of the Magnifier architecture, we compare the ``single models'' with the ``magnified'' versions. The single models are trained on the original single (``coarse'') grained images in $S_L$, as commonly done in previous works~\cite{double_step,california_no_dataset}. 
On the one hand, this allows comparing the magnifier with state-of-the-art methods trained on single-grained Sentinel-2 images. On the other hand, this can be seen as an ablation study that analyzes the impact of the second branch, which is associated with the ``fine-grained''/smaller patches, on the prediction quality.
\Cref{sec:basemodels} presents the base segmentation model architectures, and \Cref{sec:backbone} introduces the encoder used in each architecture as a feature extractor.
\Cref{sec:competitors} analyzes the existing competitors, \revone{\Cref{sec:metrics} describes the metrics used for the quantitative comparison}, and \cref{sec:settings} reports the experimental settings. \Cref{sec:results} provides the experimental results of the evaluation campaign. Finally, \Cref{sec:transfer} reports a transfer learning analysis between different areas.

\begin{figure}[t]
    \centering
    \subfloat[DeepLabV3+ \cite{deeplabv3}]{\includegraphics[width=\linewidth]{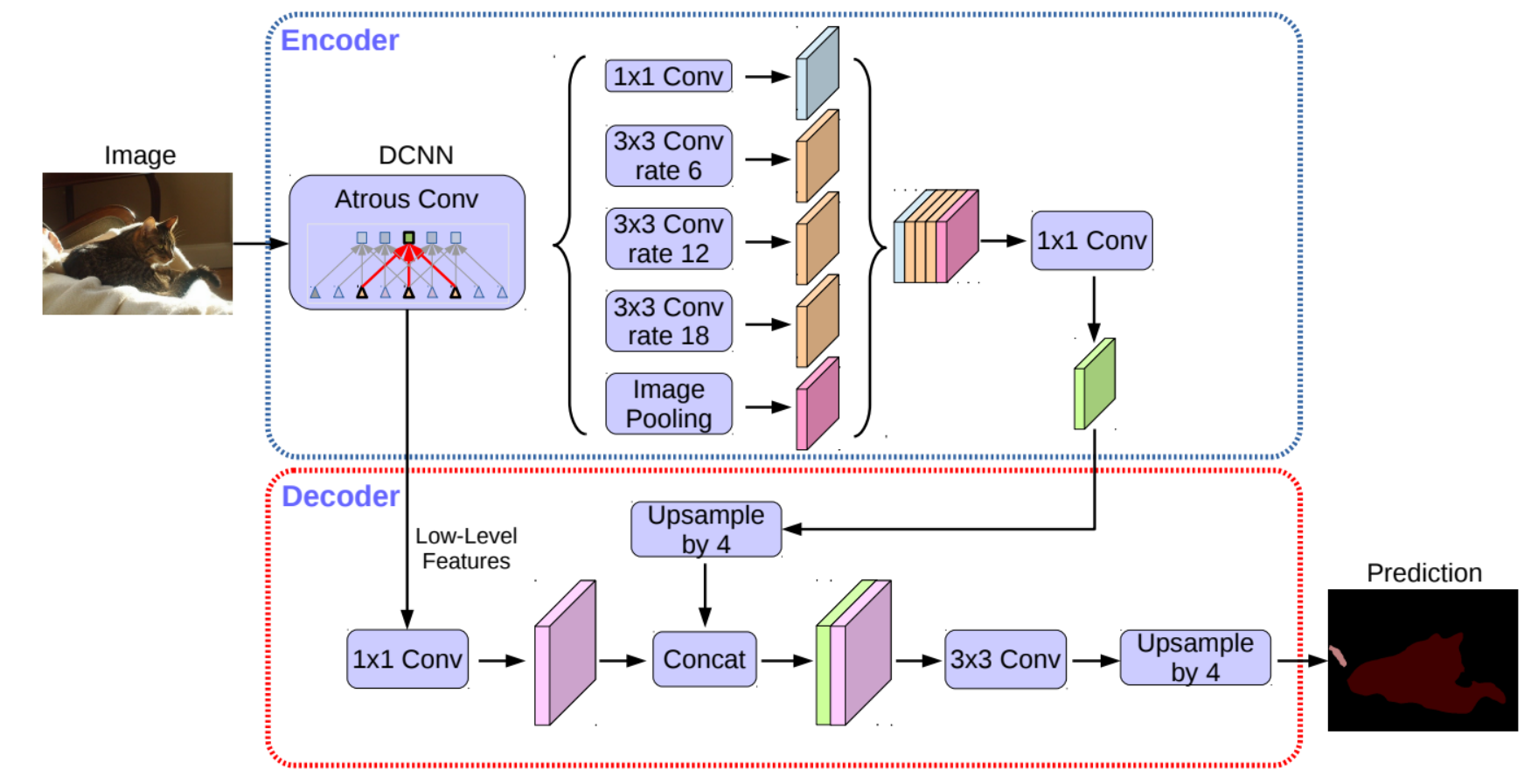}} \hfil \\
    \subfloat[U-Net \cite{unet}]{\includegraphics[width=\linewidth]{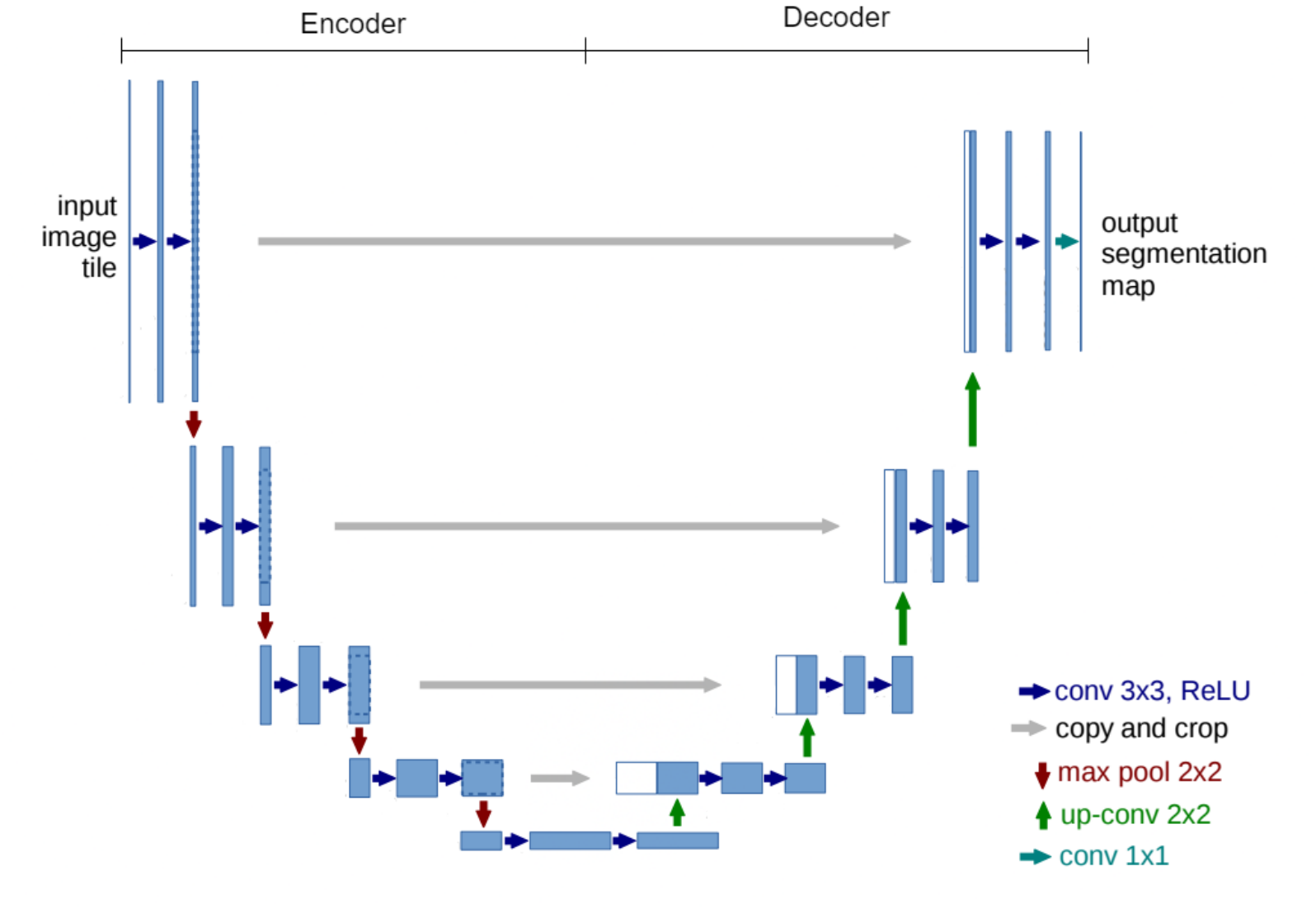}} \hfil \\
    \subfloat[SegFormer \cite{segformer}]{\includegraphics[width=\linewidth]{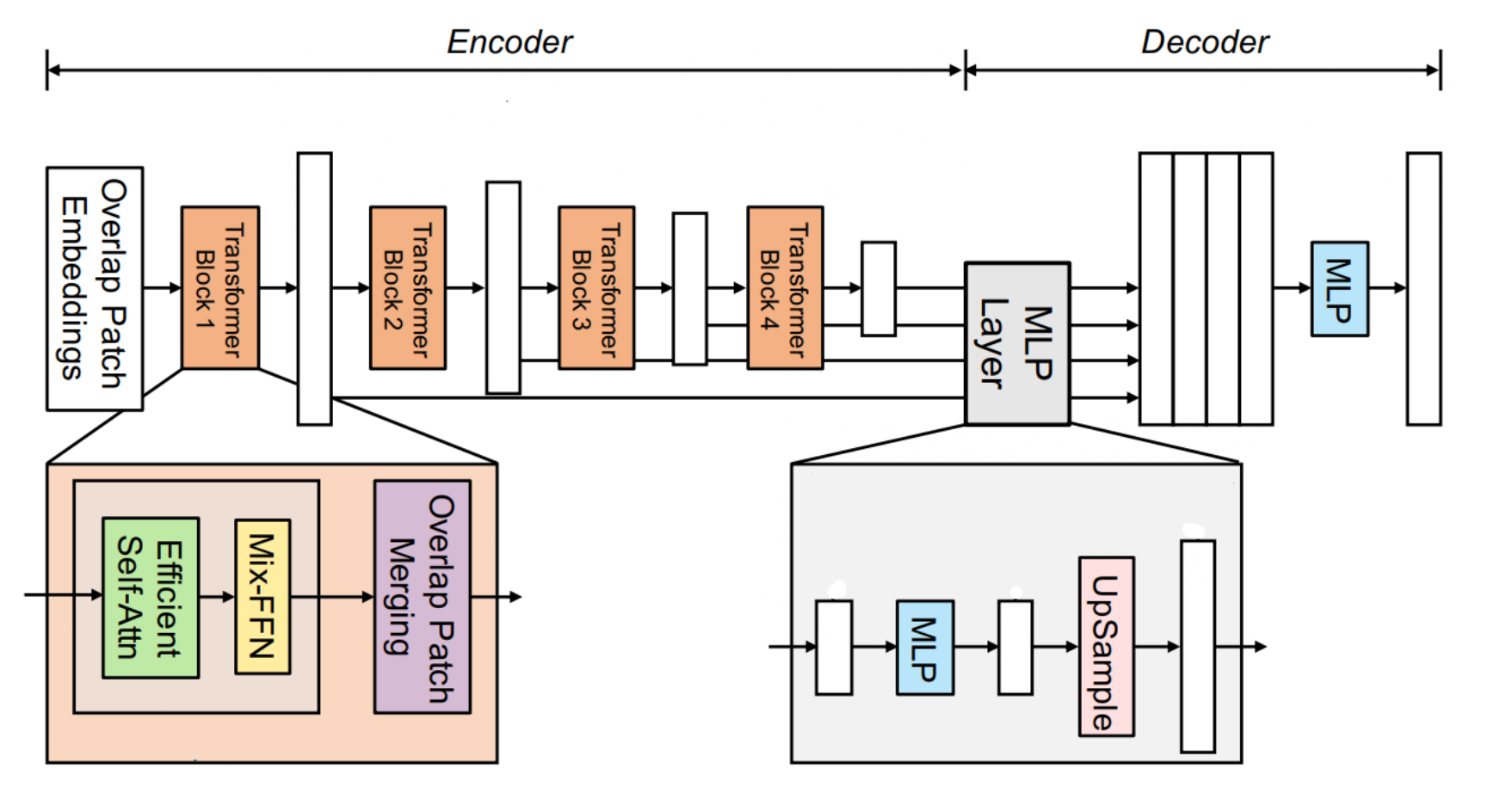}}
    \caption{Base architectures of DeepLabV3+, U-Net and SegFormer.}
    \label{fig:architectures}
\end{figure}

\subsection{Base model architectures}\label{sec:basemodels}
The chosen base model architectures, as highlighted in \Cref{fig:architectures}, are different in their structure and technological choices. However, they are all based on two distinct, non-overlapping parts: an encoder and a decoder. We selected three of the most effective and well-known semantic segmentation architectures. 
Brief descriptions of these architectures follow.

\smallskip
\subsubsection{DeepLabV3+}
The encoder module of DeepLabV3+~\cite{deeplabv3plus} incorporates multi-scale contextual information using the spatial pyramid pooling module and atrous separable convolution, a combination of atrous convolution and depthwise separable convolution (see~\Cref{fig:architectures}a). The decoder module is designed to recover lost spatial information during encoding. 

\smallskip
\subsubsection{U-Net}
The encoder of U-Net~\cite{unet}, called the contracting path, reduces the dimensionality of the input. In contrast, the decoder, called the expansive path, gradually increases the dimensionality of the feature maps (see~\Cref{fig:architectures}b). Using skip connections, the encoder extracts features of different spatial resolutions concatenated with the embeddings at the same level in the decoder.

\smallskip
\subsubsection{SegFormer}
The encoder of SegFormer~\cite{segformer} is positional-encoding-free, hierarchical, and composed of transformer blocks, which output multiscale features. The decoder comprises MLPs that aggregate local and global attention information to predict the final mask (see~\Cref{fig:architectures}c).

\smallskip
These are three of the best-performing and most frequently used models for solving the semantic segmentation problem, including the burned area delineation/segmentation task.

For DeepLabV3+ and U-Net architectures, we evaluated two different backbones designed for devices with different resources (see \Cref{sec:backbone}). Instead, for the SegFormer model, we considered the standard encoder~\cite{segformer} proposed in the original paper since it is already adaptable to different resource constraints.

\subsection{Backbone selection}
\label{sec:backbone}
In our experiments, we have chosen well-known backbones available in different ``versions.'' This means all model versions share the same architecture and only differ in the number of parameters. 
We selected two versions for each backbone. They are referred to as ``small'' and ``large'' versions in the following. The Magnifier architecture has been applied only to the small models. The large models are used only to compare the Magnifier with larger single models. 

In \Cref{tab:comparison,fig:iou_vs_size}, it is possible to see the number of parameters for the various backbones. As expected, since Magnifier is based on two branches and hence two backbones, its encoders always have a number of parameters that are twice the parameters of the small single models on top of which it is applied. However, each Magnifier model is always characterized by fewer parameters than the ``large'' version of the corresponding single model (approximately 50\% of the parameters).
The head size is not reported because all models in a family share the same head.

We applied two well-known families of backbones to DeepLabV3+ and U-Net (the two CNN-based architectures we considered). Specifically, \textbf{MobileNetV3}~\cite{mobilenet}, a state-of-art backbone for low-resource devices, and \textbf{ResNet}~\cite{resnet}, which is a state-of-art solution for many tasks. 
We evaluated a single transformer-based backbone, \textbf{MiT}, because it is easy to adapt, has many different versions, and is the one proposed in the SOTA SegFormer transformer. 

\smallskip
\subsubsection{MobileNetV3}
Mobile Nets are encoders designed to work in low-resource devices. They use Squeeze-and-Excite blocks, depthwise separable convolution, and hard-swish to reduce the computational cost without decreasing the global accuracy. The introduction of Squeeze-and-Excite permits the emulation of the attention layers at a lower cost by applying average pooling and two convolutions. The encoder is mainly composed of 3x3 and 5x5 bottleneck layers \cite{mobilenet}.

\smallskip
\subsubsection{ResNet}
ResNet is designed to overcome the issues regarding exploding and vanishing gradients. Adding shortcut connections enables computed features to reach deeper levels of the network. This permits the creation of very deep networks without losing training stability and increasing the model's accuracy. The network mainly comprises 3x3 convolutional layers, bottleneck layers, and shortcut connections~\cite{resnet}.

\smallskip
\subsubsection{MiT}
Mix Transformer (MiT) is the hierarchical encoder proposed in SegFormer. It is composed of Transformer blocks, each one containing Self-Attention, Mix-Feed Forward Network (Mix-FNN), and Overlap Patch Merging (used to generate CNN-like multi-level embeddings providing high-resolution coarse features and low-resolution fine-grained features). This encoder combines convolution and MLP of the Mix-FFN with the mechanism of Self-Attention, which is formulated with particular consideration to the efficiency~\cite{segformer}.

\subsection{Competitors}
\label{sec:competitors}
The current state-of-the-art deep learning-based models for burned area delineation using satellite images are based on ``single-grained'' images associated with a traditional semantic segmentation model (e.g., 
\cite{farasin2020supervised, ColombaCikm22}). Hence, as competitors, we have considered single state-of-the-art deep-learning models for semantic segmentation (both CNNs and vision transformers).
We also included index-based methodologies with the Otsu segmentation method \cite{otsu1979threshold}. We selected a custom state-of-the-art architecture (BurnNet) as the baseline algorithm to compare Magnifier's performances.

\revone{\subsection{Evaluation metrics}\label{sec:metrics}}
We report weighted mean F1 Score and IoU (intersection over Union) for the positive class (burned area) and the relative standard deviation computed across the folds since the number of samples per fold varies significantly according to the original settings ($\approx 64 \pm 32$ for Europe and $\approx 106 \pm 13$ for California). The F1-Score is more adequate for imbalanced classes, while IoU provides a good segmentation quality assessment. They are formulated as follows:
\begin{subequations}
    \begin{align}
        F1 = \frac{2TP}{2TP + FP + FN} \\
        IoU = \frac{TP}{TP + FP + FN}        
    \end{align}
\end{subequations}
where TP, TN, FP, and FN are the number of True Positive, True Negative, False Positive, and False Negative predictions, respectively. 

\revgeneral{
We also report the Mean Rank (MR) for each algorithm (model) across the datasets to better evaluate the differences between the analyzed methods \cite{janez2006}. MR is formulated as follows:
\begin{equation}
    MR = \frac 1 {md} \sum_{i=1}^m\sum_{j=1}^{d} r_{ji}
\end{equation}
where $r_{ji}$ is the rank (where 1 is assigned to the best) of the algorithm (model) among all algorithms for a given metric $i$ on a dataset $j$, $m$ is the total number of metrics, and $d$ is the number of datasets. MR ranges from 1 to the number of compared algorithms/models. The smaller the MR value, the better the algorithm/model. 
}

We evaluate the resource consumption of the networks expressed in GFLOPs. 
GFLOPs measure the number of mathematical operations a system is capable of performing per second and are often used to evaluate the computational demands of networks. A higher number of GFLOPs indicates a more complex model that requires more resources, such as CPU or GPU, to be processed and trained effectively. 

\subsection{Experimental settings}
\label{sec:settings}
Since the pretrains are not typical for such a type of image (with 12 or 8 channels), we randomly initialized the networks.
We used the AdamW optimizer and a polynomial learning rate scheduler with a power of 1 for 55 iterations. SegFormers are trained with a starting learning rate of 0.001, UNets with 0.0001, ResNet-DeepLabV3+s with 0.01, and Mobile-DeepLabV3+s with 0.0001. According to the original formulation, the hyper-parameters of AUF loss are $\lambda = 0.5, \delta = 0.6, \gamma = 0.1$. We kept the same configuration of the learning rate scheduler and AUF loss for BurntNet with a learning rate of 0.0001.
A patch size of $64 \times 64$ was adopted for the smaller crops in Magnifier. All models were trained on a single Tesla V100 32GB GPU. 

We applied five-fold cross-validation on the CaBuAr dataset and seven-fold cross-validation on the European one as in the original evaluation procedure~\cite{double_step,cabuar}. \revone{For the Indonesian dataset, we applied the five-fold cross-validation approach since no settings or train-test splitting were provided in the original paper.}
This means for $K$ times, where $K$ is the number of folds, we choose a fold for validation and another for testing, while the remaining ones are used for the model's training \cite{cross-validation}. This statistical method provides a more robust evaluation of classifiers than the standard way of creating a single training-validation-testing split.

The solution presented in~\cite{california_no_dataset} combines many different data sources using an ensemble of U-Net models, each trained on one data source. Since we are interested in methods based only on a single source, we do not consider~\cite{california_no_dataset} in our experimental comparison.

\subsection{Results}
\label{sec:results}

\begin{table*}[th!]
    \centering
    \revone{
    \caption{Results for the different models on California and Europe datasets. Each table refers to a specific architecture, and results are grouped by the backbone type and sorted by size (in terms of the number of parameters \textbf{BS}). We have not reported the head size since it is constant for each model. We reported the mean rank (\textbf{MR}) for each group.}
    \label{tab:comparison}
    \subfloat[DeepLabV3+\label{tab:results_deeplab}]{\input{tables/deeplab}} \\
    \subfloat[U-Net\label{tab:results_unet}]{\input{tables/unet}} \\
    \subfloat[SegFormer\label{tab:results_segformer}]{\input{tables/segformer}}\\
    \subfloat[Indexes\label{tab:results_indexes}]{\input{tables/indexes_results}} \\
    \subfloat[Comparisons\label{tab:results_comparisons}]
    {\resizebox{\linewidth}{!}{\input{tables/comparison_best}}}
    }
\end{table*}

In \Cref{tab:comparison}, we present the results for the three considered architectures (see \Cref{sec:basemodels}) and the different backbones (see \Cref{sec:backbone}) on the \revone{three} datasets. 
We would like to remind you that we trained one Magnifier model for each combination (base model architecture,  backbone type), considering the smallest backbone version for each backbone type (i.e.; ResNet-18, MobileNetV3-Small, and MiT-B0). Hence, five different models based on the Magnifier approach have been trained for each dataset (see \Cref{tab:comparison}).

\bigskip
\subsubsection{Overall Magnifier performances}
We initially analyze the results reported in \Cref{tab:comparison} to identify the best-performing model independently of the used base model architecture and backbone. 
DeepLabV3+ with Magnifier-ResNet18 achieves the highest F1 score (77.8 on CaBuAr, 83.7 on Europe and \revgeneral{80.2 on Indonesia}) and IoU (64.0 on CaBuAr, 72.7 on Europe, \revgeneral{73.5 on Indonesia}) on all datasets (see \Cref{tab:results_deeplab}).
The improvement in terms of F1 score compared to the second-best-performing competitor is higher on the CaBuAr dataset (+1.8\% with respect to DeepLabV3+ with ResNet101)\revgeneral{ and on the Indonesia one (+1.5\% compared to DeepLabV3+ with ResNet18)}. In comparison, it is only +0.1\% with respect to DeepLabV3+ with ResNet18 on the European dataset. Similar improvements are achieved in terms of IoU.

DeepLabV3+ with Magnifier-ResNet18 is always the best-performing model, independently of the dataset and evaluation metric. Conversely, the second-best-performing model among the competitors varies for the datasets (DeepLabV3+ with ResNet101 on CaBuAr, and DeepLabV3+ with ResNet18 on Europe and Indonesia). 
DeepLabV3+ equipped with ResNet18, the smallest version of considered ResNet, probably performs better than its larger version (ResNet101)\revgeneral{ in Europe and Indonesia due to the size of the datasets}. They contain fewer samples than CaBuAr. Hence, more data may be needed to train the larger model.

Overall, the use of the combination (DeepLabV3+, ResNet), with or without the application of the Magnifier approach, is the best choice. \revgeneral{The mean rank of this model, when compared to other solutions (see \Cref{tab:results_comparisons}), confirms this claim.}

Since the MobileNetV3 backbone is smaller than ResNet, MobileNetV3 performs worse than ResNet, independently of the model size and base model architecture.

The SegFormer-based models achieve intermediate results compared to the DeepLabeV3+-based and U-Net-based ones. 

\bigskip
\subsubsection{Neural Networks vs. Index-based segmentation}
Otsu's automatic thresholding applied to several indexes performs poorly in the burned area delineation problem. We validated each index on a per-fold basis despite not requiring any training procedures, such that the final scores can be compared against the ML-based methodologies.\revgeneral{ BAIS2 was not applied to the Indonesia dataset since it exploits spectral bands available in Sentinel-2.
NBR2 index achieved the overall best results on all datasets with a mean rank of 1, followed by NBR (\Cref{tab:results_indexes}). As can be seen from~\Cref{tab:results_deeplab} and~\Cref{tab:results_unet}, MobileNetV3-based variants of DeepLab and U-Net are the worst-performing deep learning models on all datasets. However, those deep learning models also outperform by a large margin the traditional index-based methodology, motivating the use of deep neural networks.}

\bigskip
\subsubsection{Magnifier vs. traditional deep models}
\label{sec:magnifiervssmallmodels}

We now analyze the improvement of applying the Magnifier approach on top of traditional models to understand its impact. We recall that the smallest version of each conventional backbone (i.e.; ResNet-18, MobileNetv3-Small, and MiT-B0) is used as an encoder in the two branches of the Magnifier.

When using DeepLabV3+ with the MobileNetV3-S backbone, Magnifier boosts the F1 score and IoU performances on all datasets (see the first two lines of \Cref{tab:results_deeplab}). The F1 score increases from 64.8 to 69.7 on CaBuAr (+4.9\%), from 72.6 to 79.7 on Europe (+7.1\%),\revgeneral{ and from 73.3 to 80.2 on Indonesia (+6.9\%)}. Slightly higher improvements are achieved regarding the IoU metric (+5.7\% on CaBuAr, +8.4\% on Europe\revgeneral{ and 11.5\% on Indonesia).}

The improvements are similar when using the DeepLabV3+ with the ResNet18 backbone applied to the CaBuAr dataset (+4.0\% in terms of F1 score and +4.6\% in IoU). Conversely, the improvements given by Magnifier are close to zero on the European dataset.\revgeneral{ In Indonesia, we still get +1.0\% in F1 and +1.5\% in IoU.}

Looking at the combination (U-Net, MobileNetV3-S) in \Cref{tab:results_unet}, Magnifier confirms its boost compared to the single model (U-Net with MobileNetV3-S) on all metrics and datasets (+4.7\% in terms F1 score and +4.6\% in IoU on CaBuAr, +3.2\% in F1 score and +4.1\% in IoU on Europe, and +7.8\% in F1 and +10.5\% in IoU on Indonesia). In this case, similarly to the previous configuration, the improvement given by using Magnifier on the CaBuAr\revgeneral{ and Indonesia datasets is slightly less evident when considering the ResNet18 backbone and worsens on the European dataset. However, it is still the best solution according to the mean rank (1.5 compared to 2 of ResNet-18 alone).}

Regarding the results obtained with SegFormer (see \Cref{tab:results_segformer}), Magnifier performs slightly better than the single model (SegFormer with MiT-B0) on Europe (+0.7\% in terms of F1 score and +0.6\% in IoU).  On the CaBuAr dataset, the Magnifier improves only the IoU metric (+0.4\%).\revgeneral{ In Indonesia, Segformer-B1 remains the best-performing solution.}

\revgeneral{The use of the Magnifier approach proves effective, as summarized by the mean ranks.} Magnifier gets the best mean rank in 4 out of 5 configurations (\Cref{tab:comparison}). \revtwo{In the single failure case in terms of MR (the one based on SegFormer - \Cref{tab:results_segformer}), all the configurations achieve similar ranks, probably due to the limitations of the transformer backbone itself.}

The reported results confirm the positive impact of the Magnifier approach compared to single models. Without increasing the number of labeled data, the combination of multi-grained versions of the images introduced in Magnifier boosts the F1 and IoU metrics on average. Moreover, in addition to the advantage in terms of segmentation performance, Magnifier shows superiority also in terms of the amount of computational resources, with lower values of FLOPs compared to the enhanced, larger versions of each model. The only exception is observed for the U-Net model and MobileNetV3 backbone, where the Magnifier model shows a slightly higher value of a number of operations. This is mainly due to the high optimization of the MobileNetV3 model and the low amount of layers being present: the duplication of the encoder component rivals the number of parameters of the larger version.

The experiments described in this section can also be seen as an ablation study in which only the first path of the Magnifier is used.
The results show that an additional branch with the same input images but at a different ``granularity'' can boost the model's performance. 

\bigskip
\subsubsection{Magnifier vs. large single models}
We finally performed experiments training single models based on larger versions of the three considered backbones (MobileNetV3-L, ResNet101, and MiT-B1). This experiment aims to understand if the improvements achieved by Magnifier compared to the smaller models (see \Cref{sec:magnifiervssmallmodels}) are due to the increase in the number of parameters or to the use of images of different sizes and granularities (i.e., to the use of the Magnifier approach).
The large backbones considered in these experiments are characterized by approximately twice the parameters of the Magnifier models and four times those of the small single models.

Considering all the combinations (base model architecture, large backbone) and all datasets, Magnifier, trained on the small versions of the backbones, performs better than the large single models in terms of F1 score\revgeneral{ 14 times out of 15 cases. In comparison, it performs better than the large models in terms of IoU in 11 cases over 15 (see \Cref{tab:results_deeplab,tab:results_unet,tab:results_segformer}).} The mean ranks highlight large models perform worse than smaller models in many cases: increasing the parameters is not the best solution with a limited amount of data due to overfitting. \revtwo{SegFormer-B1 is the only large model version that achieves a better mean rank compared to the small and Magnifier MiT-based models, but the mean ranks are quite similar, probably due to the large amount of data that transformers need to learn relations \cite{vit}. However, we recall that the MiT-based models are not the best ones overall.} 

These results support the claim that the improvement provided by Magnifier is related to its two paths and images analyzed at different levels and not to the number of parameters of the models. 
The innovative usage of the labeled data improves the quality of the trained models.

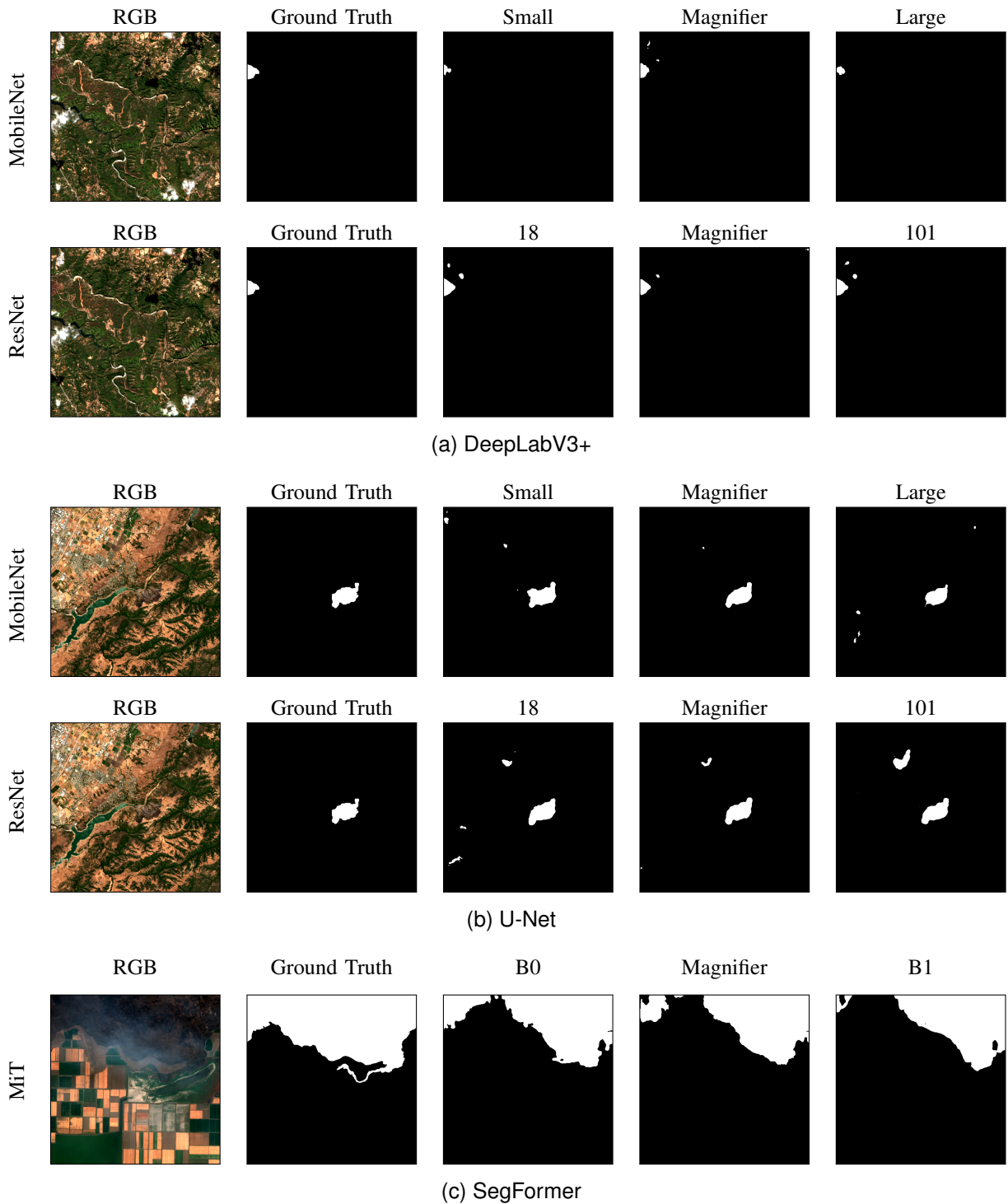
\begin{figure*}[!hbt]
    \centering
    \revone{
        \subfloat[DeepLabV3+]{\input{tables/examples_deeplab}} \\
        \subfloat[U-Net]{\input{tables/examples_unet}} \\
        \subfloat[SegFormer]{\input{tables/examples_segformer}}
        \caption{Example RGB images and corresponding ground truth with predictions. Images are grouped by architecture and type of backbone.}
        \label{fig:examples}
    }
\end{figure*}

\bigskip
\subsubsection{Magnifier vs. BurntNet}
\Cref{tab:results_comparisons} summarizes the results achieved by the overall best models previously presented, as well as the performance achieved by the BurntNet segmentation model. Focusing on the California dataset and the machine learning-based approaches, it can be observed that BurntNet achieves the overall worst performance, surpassing only SegFormer B0 with Magnifier by a low margin. Comparing the backbone and the total size, the SegFormer-based Magnifier architecture has less than half the number of trainable parameters and almost a fifth of the total number of parameters compared to BurntNet, not justifying the higher complexity of the latter. The highest performance gap in F1 and IoU scores observed in this set of experiments is \revgeneral{+6.2 and +1.8 points.} 

Instead, focusing on the Europe dataset, BurntNet showed higher performance compared to all Magnifier-enhanced models, with the largest difference being \revgeneral{+3.4 and +4.5} points in F1 and IoU scores, respectively. Considering the average results, Magnifier applied to DeepLabV3+ with ResNet18 backbone achieved the overall best result in both the evaluation metrics. The performance gap on the second dataset can be justified by (i) the larger overall dimension of the ResNet18 backbone, which might lead to overfitting issues, and (ii) the higher difficulty due to transfer learning since the seven folds adopted in the cross-validation process are region-based. In fact, the SegFormer-based model is able to surpass the U-Net model despite the lower number of parameters. Moreover, the adoption of dilation and erosion mechanisms in the BurntNet architecture enables noise removal and better shape extraction in a context in which the input dataset is smaller and noisier compared to the California dataset. 

When looking at Indonesia, BurntNet collapsed, probably due to the few samples in the dataset.

Overall, average scores demonstrate the better results of the Magnifier applied to DeepLabV3+ architecture, as well as better stability during the training procedure against BurntNet, which showed a model collapse in a few of the experimental runs, leading to invalid predictions.

Analyzing instead the number of floating point operations (GFLOPs), it can be seen that BurntNet is the least performing model, with 219 GFLOPs compared to the 76.9 GFLOPs of Magnifier DeepLabV3+ with ResNet18 backbone, with a factor of 2.84 of difference. Such difference increases if the SegFormer model is being considered, with a difference factor of 10.28. Thus, all Magnifier models are associated with a higher throughput, which is important when analyzing a large number of images or images that cover huge areas, such as in the context of Earth Observation. The erosion and dilation operations show a higher computational complexity.

\bigskip
\subsubsection{Qualitative analysis on example images}
In \Cref{fig:examples}, we present some examples of predictions using the different models analyzed in this study. We selected three different representative images (one for each base model architecture).

The examples show that Magnifier is more precise than the smallest model and generally more effective than the largest one, providing fewer false positive areas and shapes more similar to the ground truth. Magnifier reduce the noise that affects the non-Magnifier models. 

Summarizing the results reported in \Cref{tab:comparison,fig:examples}, we can conclude the best IoU and F1 score on all datasets can be obtained using DeepLabV3+ with ResNet18 encoder in combination with the proposed Magnifier architecture. 
Moreover, Magnifier boosts the average performance compared to single models.

\subsection{Transfer Learning between Continents}
\label{sec:transfer}
The datasets cover different areas of the globe with different morphological characteristics, however, we perform a cross-test of the models trained on each fold of California to the "purple" fold of Europe, and we tested the counterparts of Europe on the "1" fold of California. The models show significant degradation in both cases: the ones trained in Europe get a mean IoU of 27.2, while the ones trained in California get 14.4. The usage of Magnifier architecture does not provide any benefits in this case since it achieves a mean IoU of 22.0, while without it we can achieve 20.1. The main issue is, in both cases, the necessity of a more comprehensive collection of data to cover both areas to be capable of generalizing in both continents.
\revgeneral{
We do not apply transfer learning to Indonesia since it contains data from a different satellite with different spectral bands.
}

\section{Discussion}
The introduction of the Magnifier architecture marks an advancement in the field of remote sensing, particularly in how deep learning is applied to burned area delineation. Traditionally, the field has relied on general-purpose semantic segmentation models, which, while effective, often struggled to optimize the use of limited data. Our approach is a step in addressing these challenges.

\subsection{Data Efficiency}
Magnifier architecture has the ability to enhance performance without relying on large amounts of labeled data. Data scarcity has long been a barrier in remote sensing applications, as the manual annotation of satellite images is resource-intensive and time-consuming. While a multimodal approach can be appealing by providing more features per pixel, it increases the computation costs and requires extra data, which have to be harmonized together to provide consistent information. This approach is also difficult due to the different revolution times of different satellites. Magnifier's multi-grained approach offers a solution by optimizing existing datasets and extracting fine and coarse-level features from the same images. This allows researchers and practitioners to achieve better results without the burden of large-scale data collection efforts.

\subsection{Model Scalability}
Magnifier achieves state-of-the-art performance with small models, making it suitable for applications in resource-constrained environments. Many remote sensing operations, such as those deployed on satellites or drones, require models to be computationally efficient. The scalability of the Magnifier architecture enables its deployment in these scenarios, opening up new possibilities for real-time environmental monitoring and disaster response. While many solutions focus only on improving performance, our proposal balances practicality, throughput, and model accuracy for a widespread application.

\subsection{Multi-Granularity Feature Extraction}
The introduction of multi-grained feature extraction in the Magnifier architecture sets a new direction for future research in the remote sensing domain. Traditional models operate at a single granularity, while also extracting features at different resolutions. Magnifier has demonstrated the benefits of simultaneously capturing local and global contextual information. This highlights the importance of multi-scale learning in burned areas delineation, particularly combining the network understanding at different contextual levels and not only at different resolutions. A broader exploration with different data types, such as optical, radar, and thermal imagery, and by leveraging multi-grained information across various spectral bands, future models could improve the precision and robustness of remote sensing applications in detecting and responding to environmental changes.

\subsection{Transfer Learning}
Due to the limited amount of data available and the different morphological and phenological characteristics of the areas under analysis caused by the different geographical distributions, we observed low segmentation performances when transferring the trained model from one region to another. In fact, all the trained architectures demonstrated the inability to generalize to different areas when switching from California to Europe and vice-versa. This observation raises the need to build a large- and global-scale dataset for emergency management.

\section{Conclusion}
\label{sec:conclusion}
In this article, we presented a simple but effective way to improve the performance in burned area delineation, showing that an innovative combination of images at different levels and sizes can work better than big models. %
Semantic segmentation models are mainly designed to work with RGB images, while we have to deal with 12 or 8 channels. The combined usage of global (original images) and local (smaller patches) views has proved to grant better results with more features per pixel without increasing the number of needed labeled samples. The Magnifier architecture was applied to well-known state-of-art models, from classical CNNs to new Vision Transformers, proving its versatility. 

In future research, we will investigate the design of simpler, more efficient, and effective ways to solve the burned area delineation task without the need for expensive hardware and a large amount of data.
We will also investigate the effectiveness of Magnifier in other domains and tasks in future works.

\bibliographystyle{IEEEtran}
\bibliography{IEEEabrv,references}

\begin{IEEEbiography}[{\includegraphics[width=1in,keepaspectratio,clip]{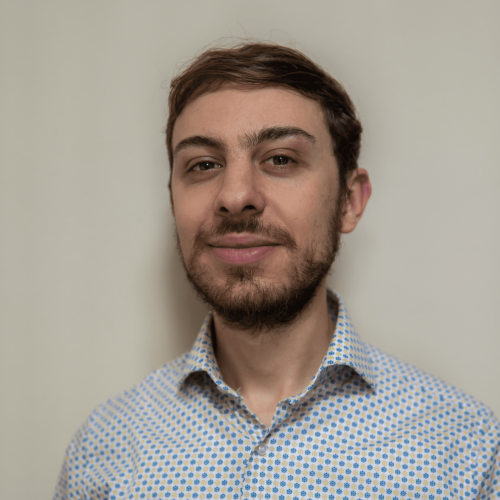}}]{Daniele Rege Cambrin}
    received a master's degree in Computer Engineering from Politecnico di Torino in 2022. He is a Ph.D. student in the same institution's Department of Control and Computer Engineering (DAUIN). His main research interests are machine learning applied to geospatial data and scalable algorithms.
\end{IEEEbiography}

\begin{IEEEbiography}[{\includegraphics[width=1in,height=1.25in,clip,keepaspectratio]{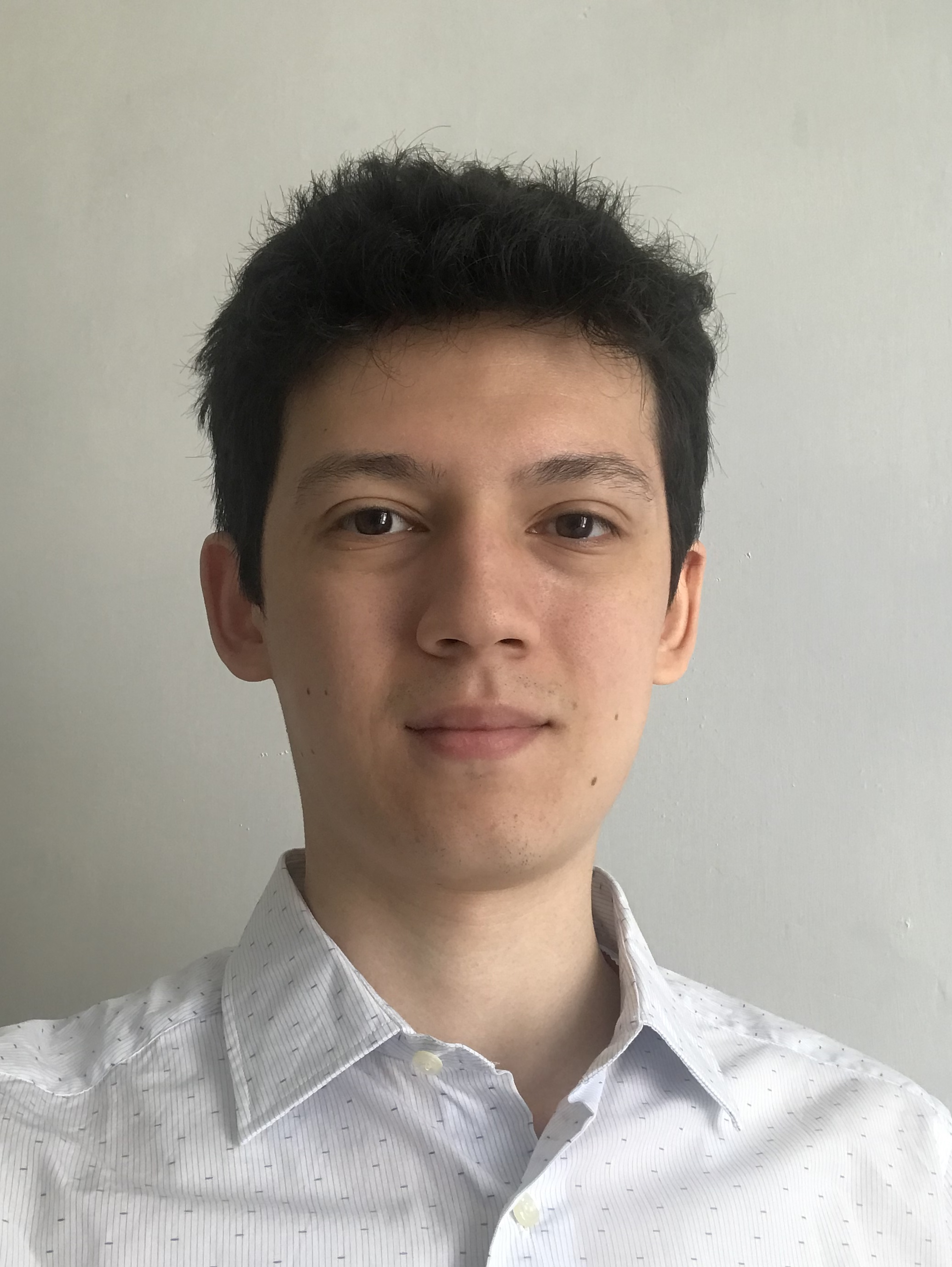}}]{Luca Colomba}
received a master's degree in Computer Engineering from Politecnico di Torino in 2019, where he is currently a Ph.D. student in the Department of Control and Computer Engineering (DAUIN). His major research interests include big data analytics, scalable algorithms, data mining, and machine learning applied to spatio-temporal data.
\end{IEEEbiography}

\begin{IEEEbiography}
    [{\includegraphics[width=1in,height=1.25in,clip,keepaspectratio]{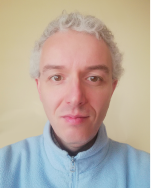}}]{Paolo Garza} received the master's and Ph.D. degrees in Computer Engineering from Politecnico di Torino in 2001 and 2005, respectively. Since December 2018, he has been an associate professor at the Dipartimento di Automatica e Informatica, Politecnico di Torino.
    His research interests include data mining algorithms, big data analytics, and data science. He has worked on classification, clustering, itemset mining, and scalable algorithms.
\end{IEEEbiography}

\end{document}

%% file: tables/s2-bands.tex
\begin{tabular}{l|c|c|l}
    \toprule
    Band & Resolution & Central wavelength & Description \\
    \midrule
    B1 &	60 m	& 443 nm	& Ultra Blue (Coastal and Aerosol) \\
    B2 &	10 m	& 490 nm	& Blue \\
    B3 &	10 m	& 560 nm	& Green \\
    B4 &	10 m	& 665 nm	& Red \\
    B5 &	20 m	& 705 nm	& Visible and Near Infrared (VNIR) \\
    B6 &	20 m	& 740 nm	& Visible and Near Infrared (VNIR) \\
    B7 &	20 m	& 783 nm	& Visible and Near Infrared (VNIR) \\
    B8 &	10 m	& 842 nm	& Visible and Near Infrared (VNIR) \\
    B8a &	20 m	& 865 nm	& Visible and Near Infrared (VNIR) \\
    B9 &	60 m	& 940 nm	& Short Wave Infrared (SWIR) \\
    B10 &	60 m	& 1375 nm	& Short Wave Infrared (SWIR) \\
    B11 &	20 m	& 1610 nm	& Short Wave Infrared (SWIR) \\
    B12 &	20 m	& 2190 nm	& Short Wave Infrared (SWIR) \\
    \bottomrule
\end{tabular}

%% file: tables/deeplab.tex
\begin{tabular}{llcc|cc|cc|cc|c}
                             &           &                     & \multicolumn{1}{l|}{} & \multicolumn{2}{c|}{California}             & \multicolumn{2}{c|}{Europe}                 & \multicolumn{2}{c|}{Indonesia}              &  \\ \midrule
Backbone type                & Backbone  & $\downarrow$ GFLOPs & $\downarrow$ BS       & F1                   & IoU                  & F1                   & IoU                  & F1                   & IoU                  & MR        \\ \midrule
\multirow{3}{*}{MobileNetV3} & Small     & 6.0                 & 0.93M                 & 64.8±7.9          & 48.4±8.5          & 72.6±15.2          & 59.1±17.6          & 73.3±6.1          & 58.1±7.6          &    2.7                     \\
                             & Magnifier & 8.6                 & 1.86M                 & \textbf{69.7±8.1} & \textbf{54.1±9.2} & \textbf{79.7±10.5} & \textbf{67.5±13.9} & \textbf{80.2±3.0}  & \textbf{69.6±4.3} & \textbf{1}  \\
                             & Large     & 9.4                 & 2.97M                 & 60.5±12.2          & 44.4±11.3          & 74.0±13.1          & 60.3±15.8          & 75.5±4.9          & 60.9±6.2          &  2.3                       \\ \midrule
\multirow{3}{*}{ResNet}      & 18        & 40.2                & 11M                   & 73.8±10.2          & 59.4±12.1          & 83.6±8.2          & \textbf{72.7±11.8} & 83.7±2.8          & 72.0±4.1          & 2.2                       \\
                             & Magnifier & 76.9                & 22M                   & \textbf{77.8±6.4} & \textbf{64.0±8.0} & \textbf{83.7±7.6} & \textbf{72.7±11.0} & \textbf{84.7±2.2} & \textbf{73.5±3.4} & \textbf{1}     \\
                             & 101       & 115.6               & 42M                   & 76.0±10.2          & 62.3±12.0          & 81.9±8.3          & 70.3±12.0          & 82.4±3.9          & 70.2±5.5          & 2.7                        \\ \bottomrule
\end{tabular}

%% file: tables/unet.tex
\begin{tabular}{@{}llcc|cc|cc|cc|c@{}}
                             &           &                     & \multicolumn{1}{l|}{} & \multicolumn{2}{c|}{California}             & \multicolumn{2}{c|}{Europe}                 & \multicolumn{2}{c|}{Indonesia}              &              \\ \midrule
Backbone type                & Backbone  & $\downarrow$ GFLOPs & $\downarrow$ BS       & F1                   & IoU                  & F1                   & IoU                  & F1                   & IoU                  & MR           \\ \midrule
\multirow{3}{*}{MobileNetV3} & Small     & 20.7                & 0.93M                 & 62.6±14.4          & 47.0±13.9          & 75.9±10.9          & 62.3±13.6          & 74.6±5.1          & 59.7±6.4          & 2.3          \\
                             & Magnifier & 25.5                & 1.86M                 & \textbf{67.3±10.9} & \textbf{51.6±11.6} & \textbf{79.1±9.6} & \textbf{66.4±12.5} & \textbf{82.4±4.0} & \textbf{70.2±5.8} & \textbf{1}   \\
                             & Large     & 24.7                & 2.97M                 & 66.3±5.8          & 49.9±6.3          & 73.5±17.2          & 60.8±19.8          & 70.6±8.5          & 55.1±10.4          & 2.7          \\ \midrule
\multirow{3}{*}{ResNet}      & 18        & 47.0                & 11M                   & 73.2±5.3          & 58.0±6.3          & \textbf{82.1±7.6} & \textbf{70.4±10.7} & 82.0±3.1          & 69.6±4.4          & 2            \\
                             & Magnifier & 78.1                & 22M                   & \textbf{74.4±5.7} & 59.6±6.9          & 81.0±9.9          & 69.2±13.3          & \textbf{82.9±4.0} & \textbf{71.0±5.8} & \textbf{1.5} \\
                             & 101       & 127.6               & 42M                   & 73.5±15.0          & \textbf{60.0±16.6} & 80.8±8.7          & 68.6±12.1          & 81.8±3.9          & 69.3±5.6          & 2.5          \\ \bottomrule
\end{tabular}

%% file: tables/segformer.tex
\begin{tabular}{@{}llcc|cc|cc|cc|c@{}}
                     &           &                     & \multicolumn{1}{l|}{} & \multicolumn{2}{c|}{California}             & \multicolumn{2}{c|}{Europe}                 & \multicolumn{2}{c|}{Indonesia}            &     \\ \midrule
Backbone type        & Backbone  & $\downarrow$ GFLOPs & $\downarrow$ BS       & F1                   & IoU                  & F1                   & IoU                  & F1                  & IoU                 & MR  \\ \midrule
\multirow{3}{*}{MiT} & B0        & 16.0                & 3M                    & \textbf{71.7±8.6} & 56.5±9.7          & 81.8±10.5          & 70.4±14.3          & 82.3±3.0          & 70.0±4.4           & 2.2 \\
                     & Magnifier & 21.3                & 6M                    & 71.5±12.0          & \textbf{56.9±13.2} & \textbf{82.5±8.0} & 71.0±11.2          & 82.2±2.4         & 69.9±3.5         & 2   \\
                     & B1        & 31.4                & 13M                   & 69.0±11.8          & 53.7±12.4          & 82.4±10.4          & \textbf{71.4±14.5} & \textbf{83.0±2.4} & \textbf{71.0±3.5} & \textbf{1.8} \\ \bottomrule
\end{tabular}

%% file: tables/indexes_results.tex
\begin{tabular}{@{}l|cc|cc|cc|c@{}}
      & \multicolumn{2}{c|}{California}             & \multicolumn{2}{c|}{Europe}                 & \multicolumn{2}{c|}{Indonesia}             &            \\ \midrule
Index & F1                   & IoU                  & F1                   & IoU                  & F1                   & IoU                 & MR         \\ \midrule
NBR   & 15.0±23.1          & 10.3±18          & 44.0±18.3          & 29.7±14.9          & 18.8±3.1          & 10.4±1.9         & 2.3        \\
NBR2  & \textbf{22.6±26.9} & \textbf{15.9±20.9} & \textbf{49.2±19.3} & \textbf{34.4±16.1} & \textbf{30.1±8.3} & \textbf{17.9±6.0} & \textbf{1} \\
BAIS  & 4.0±12.1          & 2.6±8.6          & 17.6±13.7          & 10.3±9.0          & 8.8±1.0           & 4.6±0.6         & 3.7        \\
BAIS2 & 19.4±29.2          & 14.8±25.2          & 28.5±15.9          & 17.5±10.8          & -                    & -                   & 3          \\ \bottomrule
\end{tabular}

%% file: tables/comparison_best.tex
\begin{tabular}{@{}lc|cc|cc|cc|cc|c@{}}
                          &                     & \multicolumn{2}{c|}{Number of parameters} & \multicolumn{2}{c|}{California}             & \multicolumn{2}{c|}{Europe}                 & \multicolumn{2}{c|}{Indonesia} &              \\ \midrule
Model                     & $\downarrow$ GFLOPs & $\downarrow$ BS    & $\downarrow$ Size    & F1                   & IoU                  & F1                   & IoU                  & F1             & IoU           & MR           \\ \midrule
NBR2                      & -                   & -                  & -                    & 22.6±26.9          & 15.9±20.9          & 49.2±19.3          & 34.4±16.1          & 30.1±8.3    & 17.9±6.0   & 4.7          \\
Magnifier DeepLabV3+-RN18 & 76.9                & 22M                & 24.2M                & \textbf{77.8±6.4} & \textbf{64.0±8.0} & 83.7±7.6          & 72.7±11.0          & \textbf{84.7±2.2}    & \textbf{73.5±3.4}   & \textbf{1.3} \\
Magnifier UNet-RN18       & 78.1                & 22M                & 27.5M                & 74.4±5.7          & 59.6±6.9          & 81.0±9.9          & 69.2±13.3          & 82.9±4.0    & 71.0±5.8   & 2.8          \\
Magnifier SegFormer-B0    & 21.3                & 6M                 & 7.19M                & 71.5±12.0          & 56.9±13.2          & 82.5±8.0          & 71.0±11.2          & 82.2±2.4    & 69.9±3.5   & 3.3          \\
BurntNet                  & 219.0                 & 14.5M              & 35M                  & 71.6±32.6          & 62.2±29.6          & \textbf{84.4±8.2} & \textbf{73.7±11.7} & -              & -             & 2.8          \\ \bottomrule
\end{tabular}

%% file: tables/examples_deeplab.tex
\begin{tabular}{cccccc}
& RGB
& Ground Truth    
& Small
& Magnifier
& Large
\\
\rotatebox[origin=l]{90}{\hspace{3mm} MobileNet}
&\includegraphics[width=0.15\linewidth, frame]{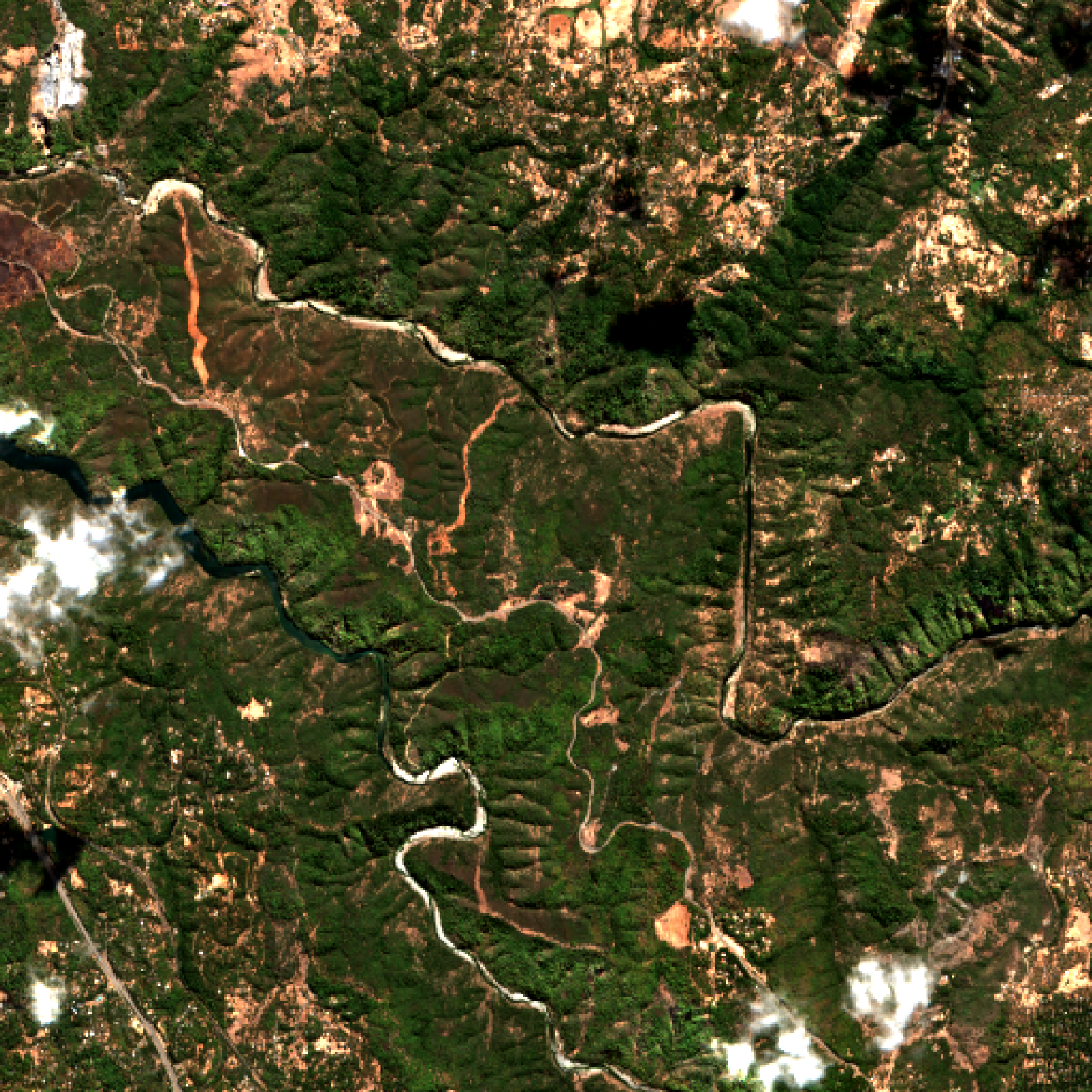}
&\includegraphics[width=0.15\linewidth, frame]{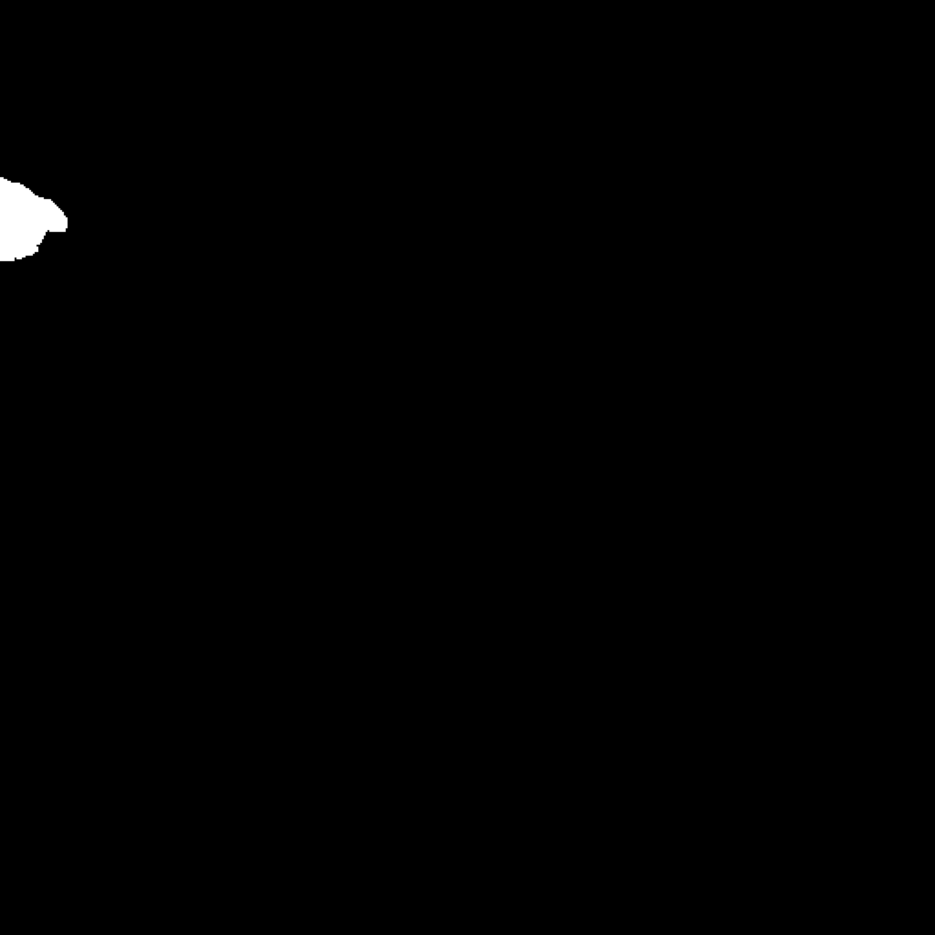}
& \includegraphics[width=0.15\linewidth, frame]{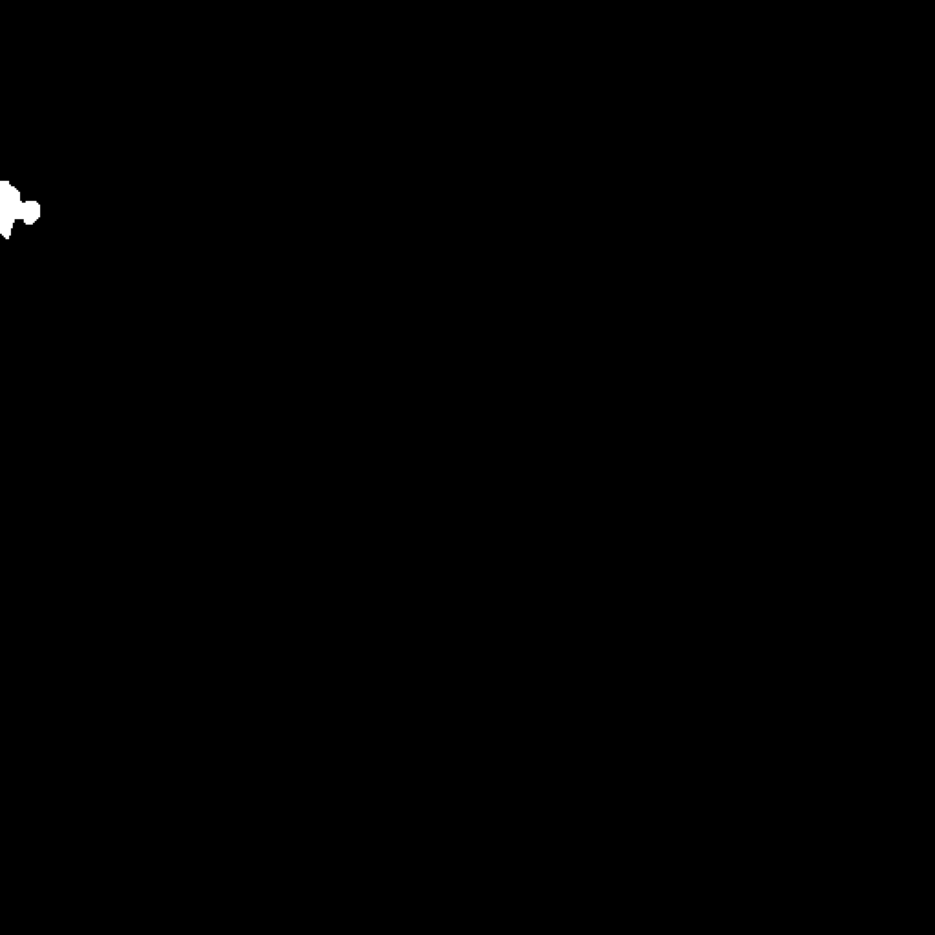}          
& \includegraphics[width=0.15\linewidth, frame]{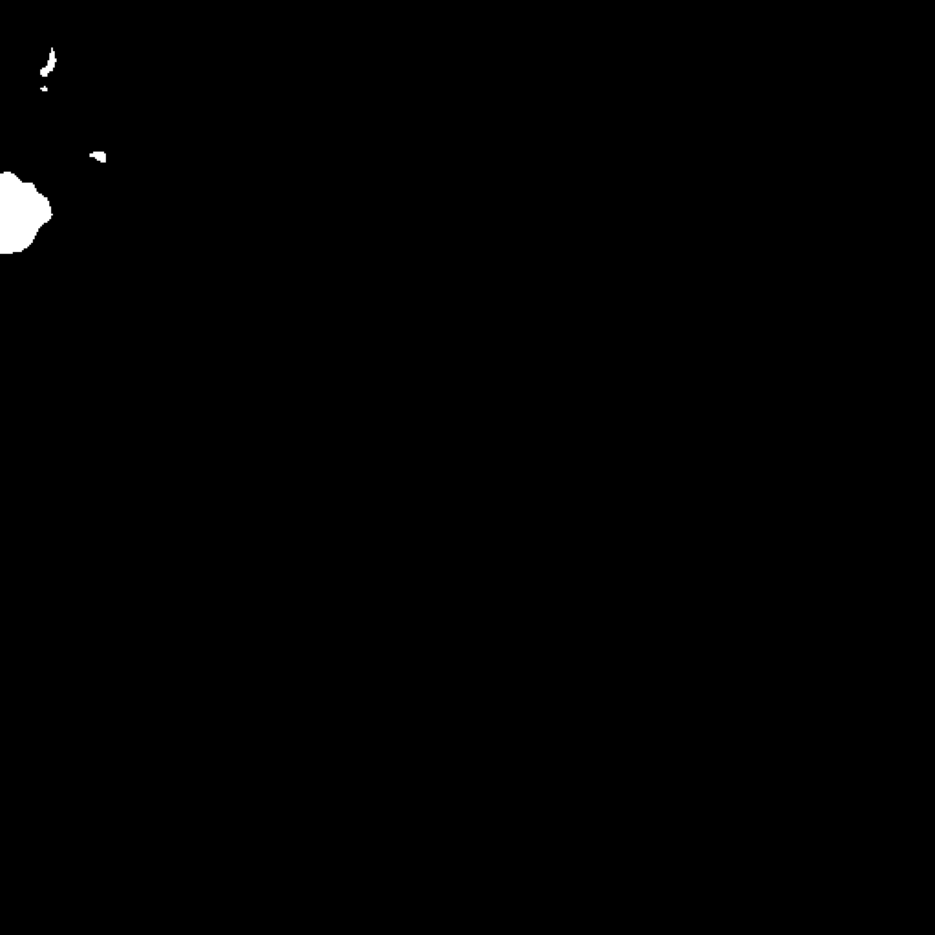}                        
& \includegraphics[width=0.15\linewidth, frame]{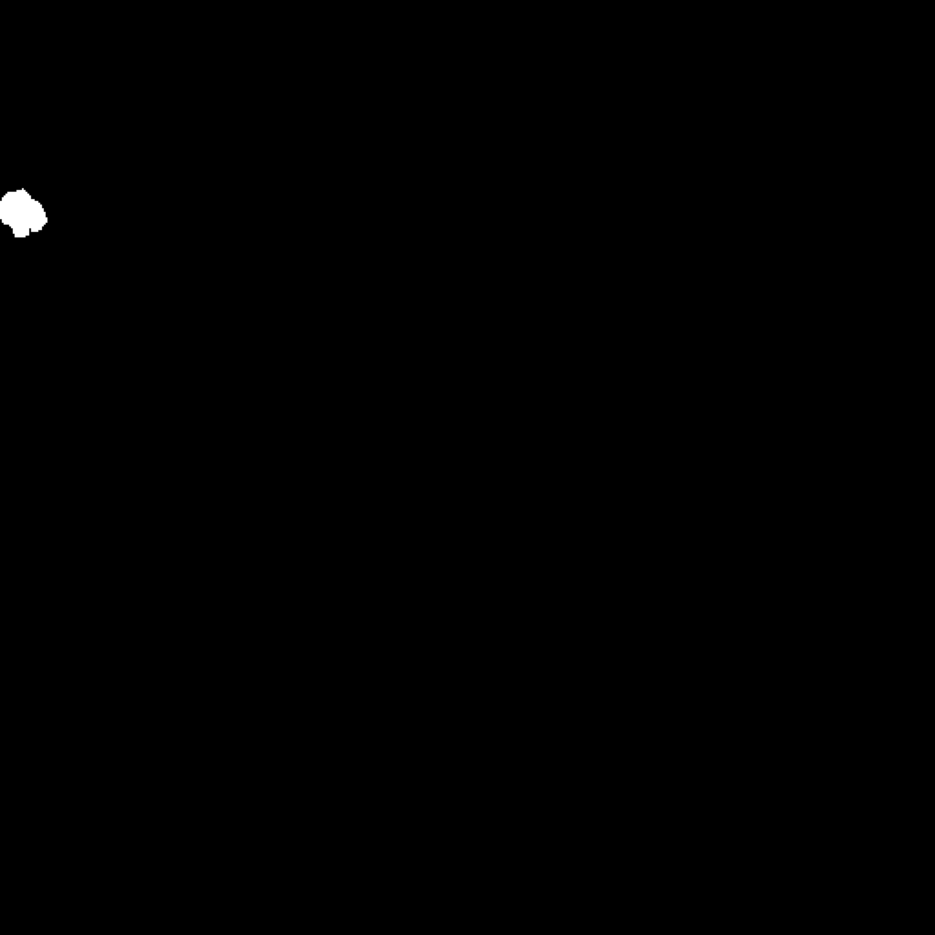}   
\\
[0.2cm]
& RGB
& Ground Truth      
& 18 
& Magnifier
& 101 
\\
 \rotatebox[origin=l]{90}{\hspace{7mm} ResNet}
& \includegraphics[width=0.15\linewidth, frame]{images/examples/resnet-deeplab/rgb.pdf}
& \includegraphics[width=0.15\linewidth, frame]{images/examples/resnet-deeplab/gt.png}
& \includegraphics[width=0.15\linewidth, frame]{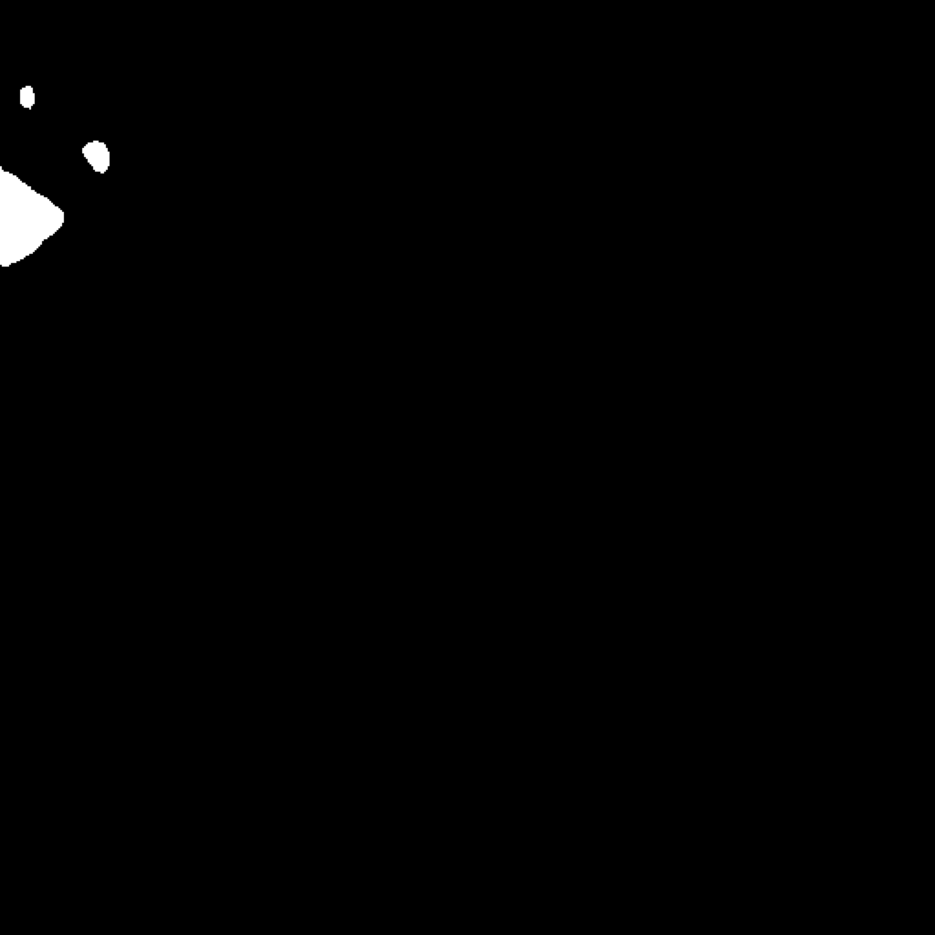}          
& \includegraphics[width=0.15\linewidth, frame]{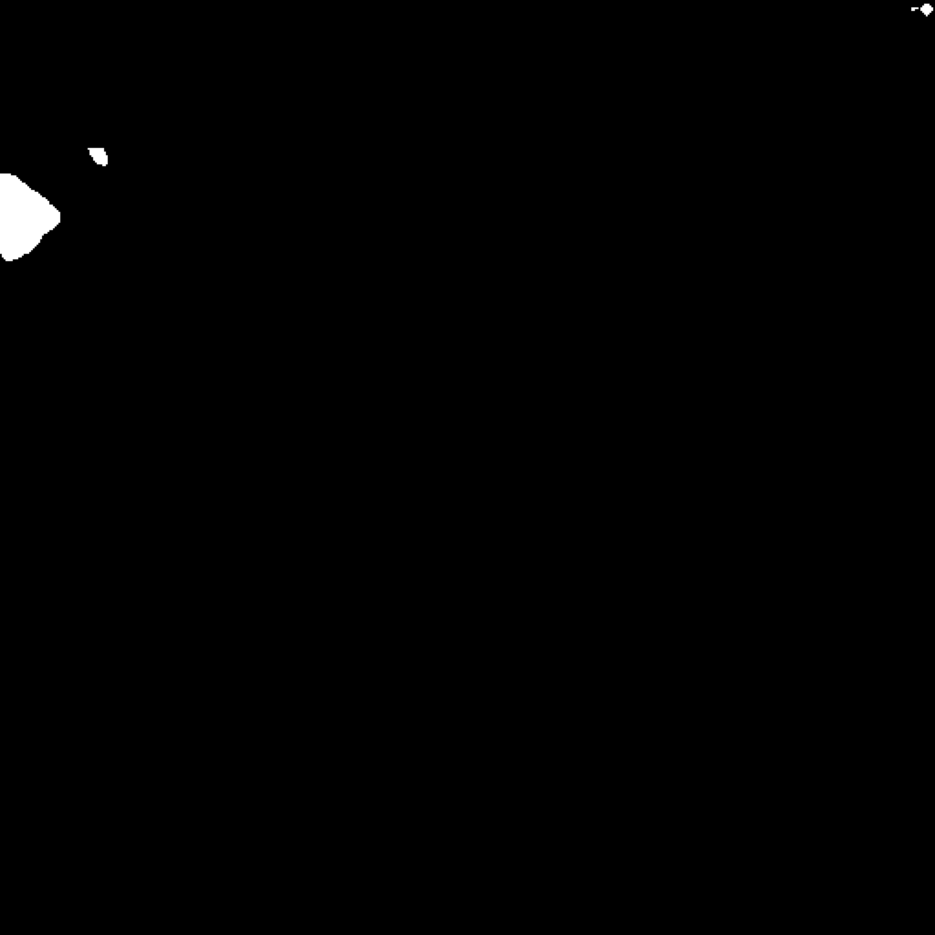}                        
& \includegraphics[width=0.15\linewidth, frame]{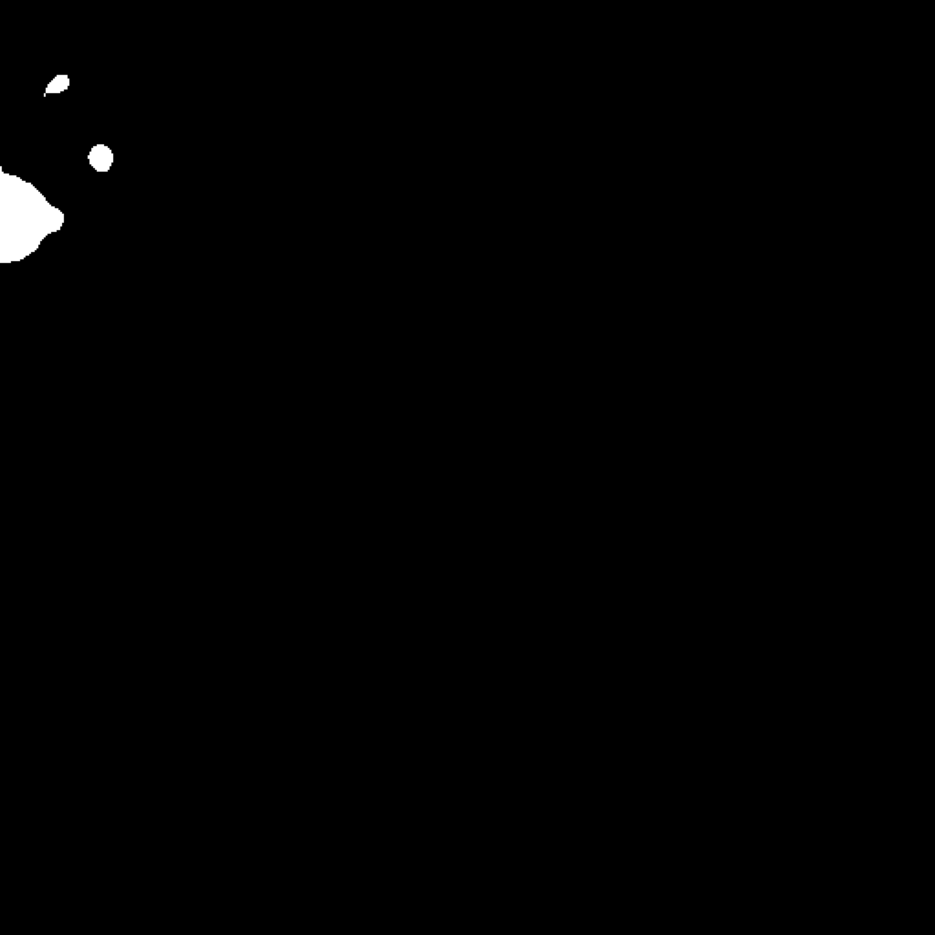}  
\\
\end{tabular}

%% file: tables/examples_unet.tex
\begin{tabular}{cccccc}
& RGB
& Ground Truth      
& Small
& Magnifier
& Large
\\
\rotatebox[origin=l]{90}{\hspace{3mm} MobileNet}
& \includegraphics[width=0.15\linewidth, frame]{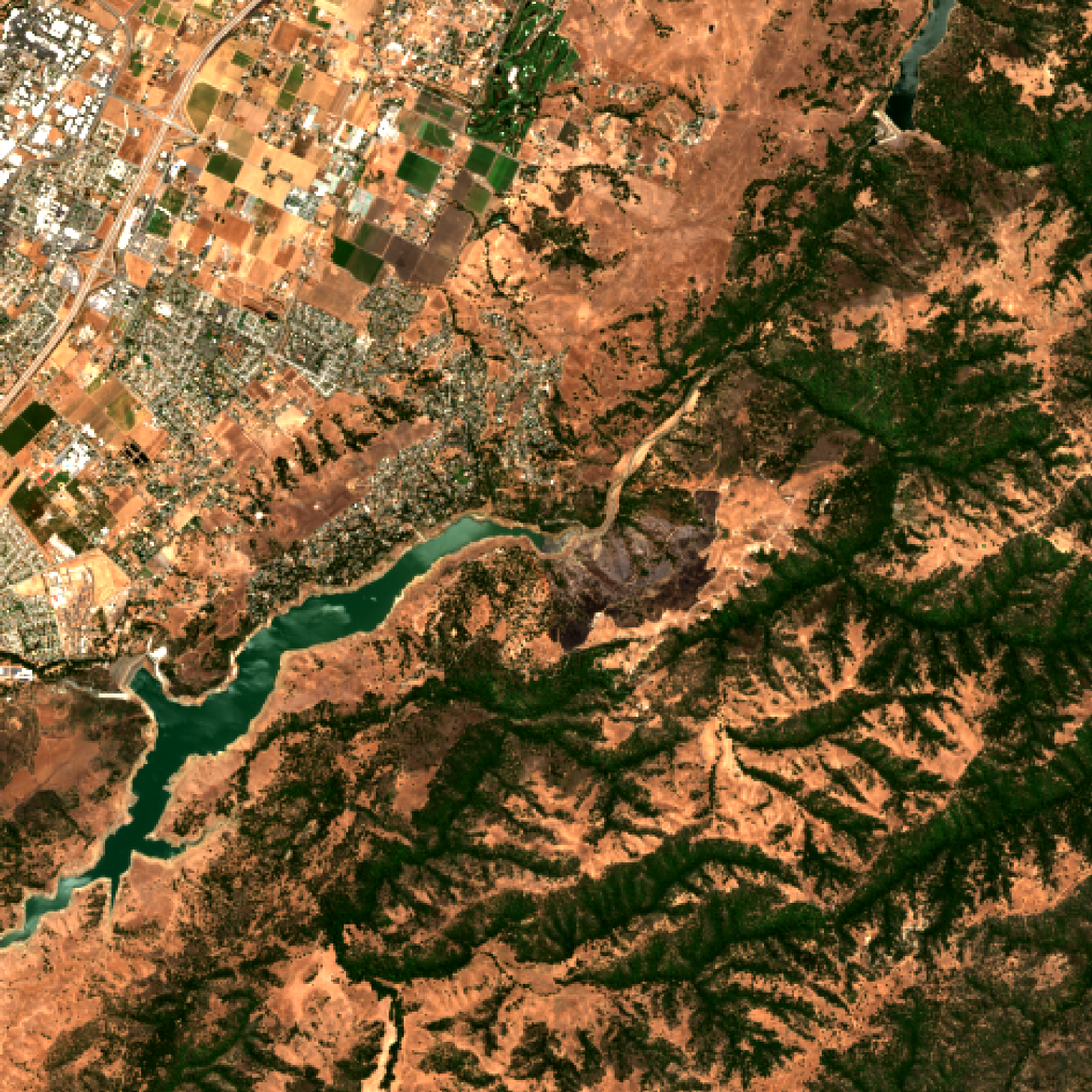}
&\includegraphics[width=0.15\linewidth, frame]{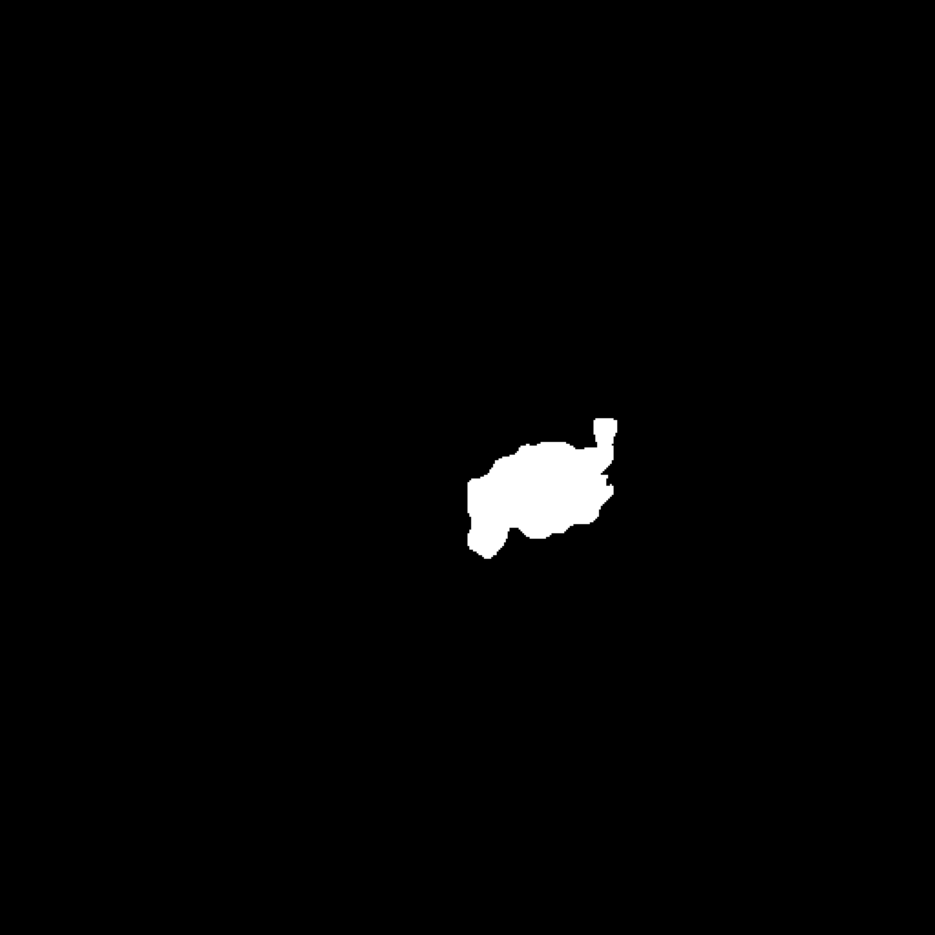}
& \includegraphics[width=0.15\linewidth, frame]{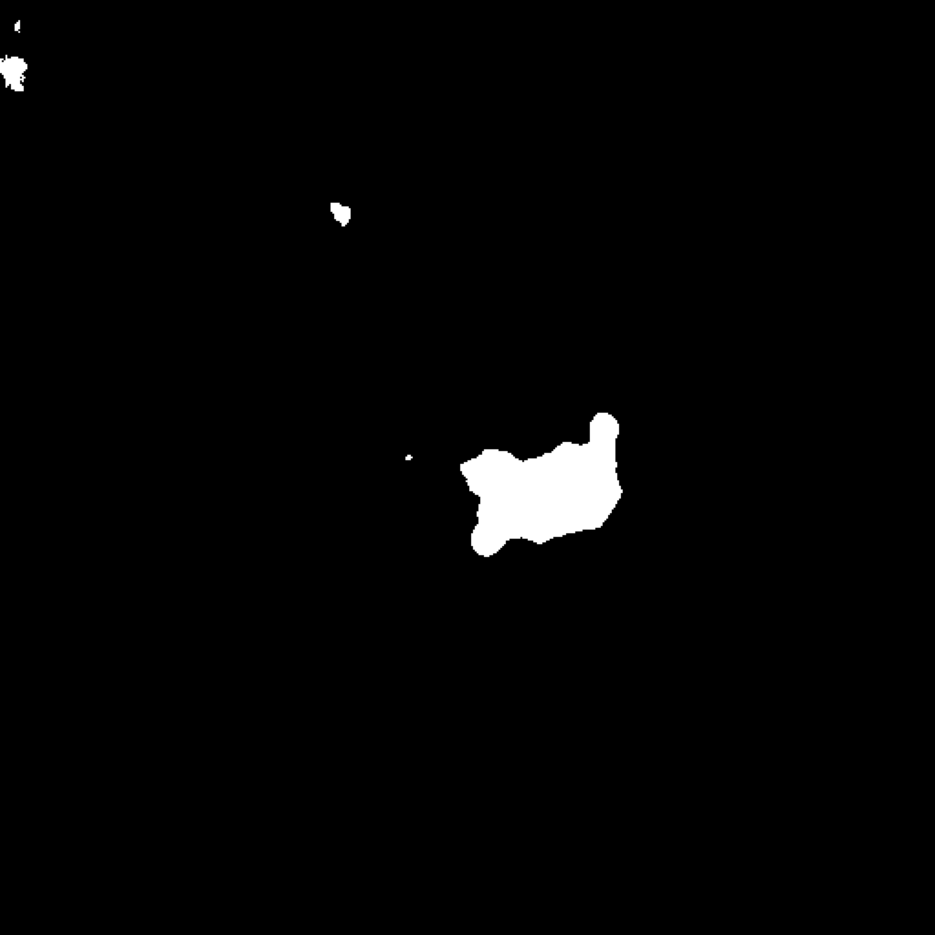}          
& \includegraphics[width=0.15\linewidth, frame]{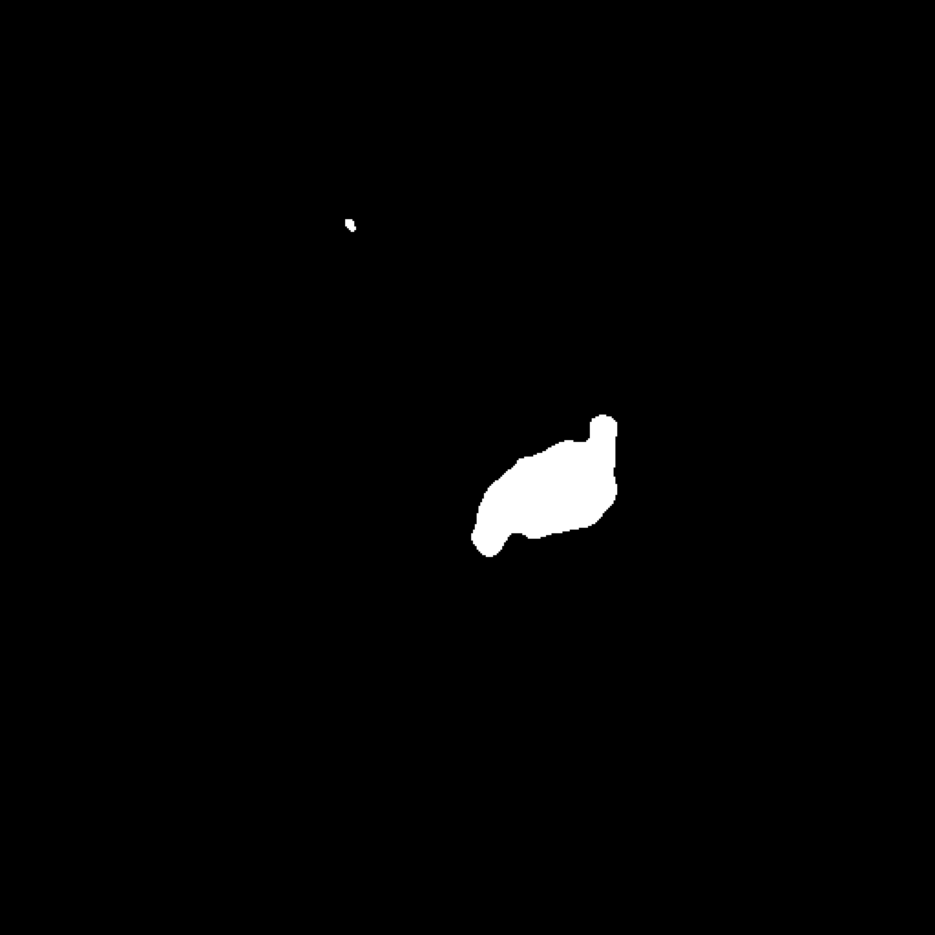}                        
& \includegraphics[width=0.15\linewidth, frame]{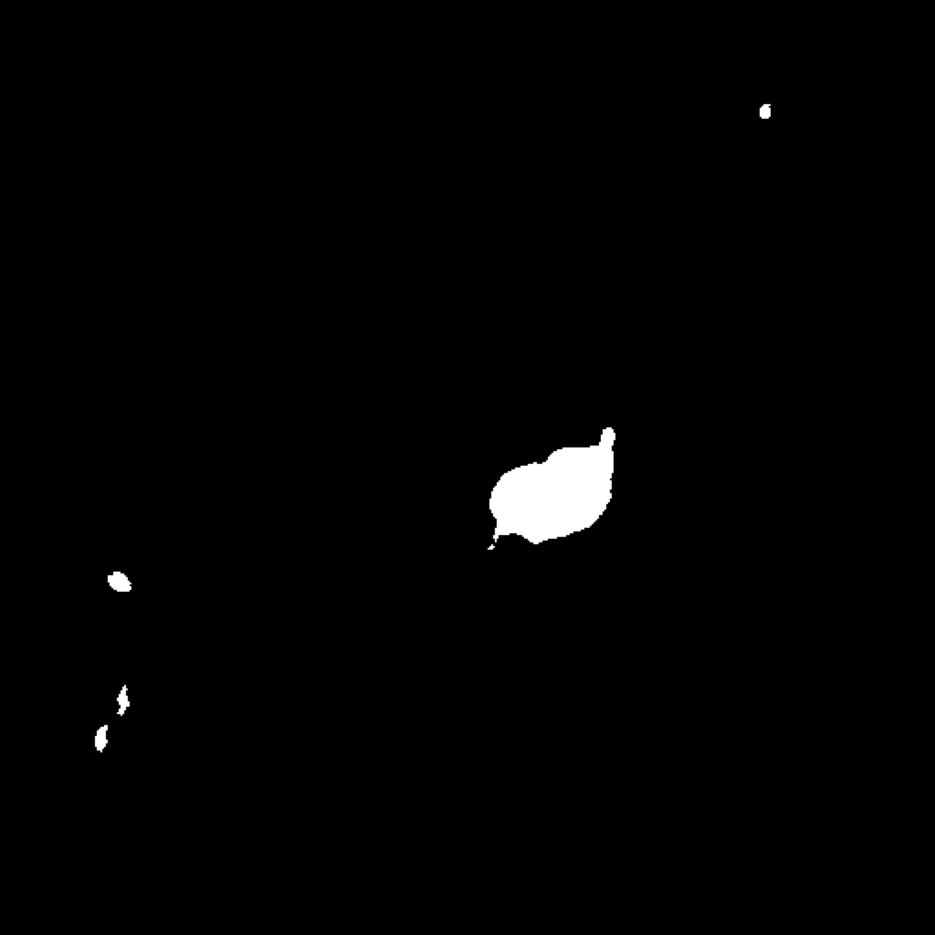}   
\\[0.2cm]
& RGB
& Ground Truth   
& 18 
& Magnifier
& 101 
\\
\rotatebox[origin=l]{90}{\hspace{7mm} ResNet}
& \includegraphics[width=0.15\linewidth, frame]{images/examples/resnet-unet/rgb.pdf}
& \includegraphics[width=0.15\linewidth, frame]{images/examples/resnet-unet/gt.png}
& \includegraphics[width=0.15\linewidth, frame]{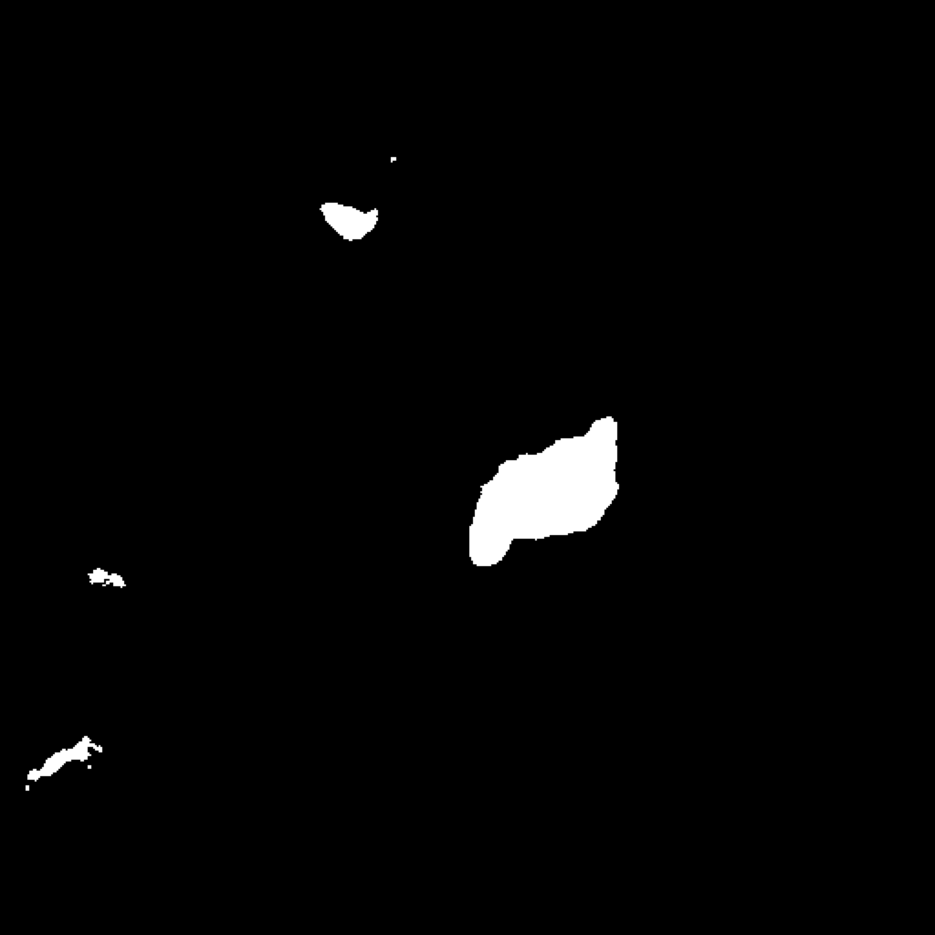}          
& \includegraphics[width=0.15\linewidth, frame]{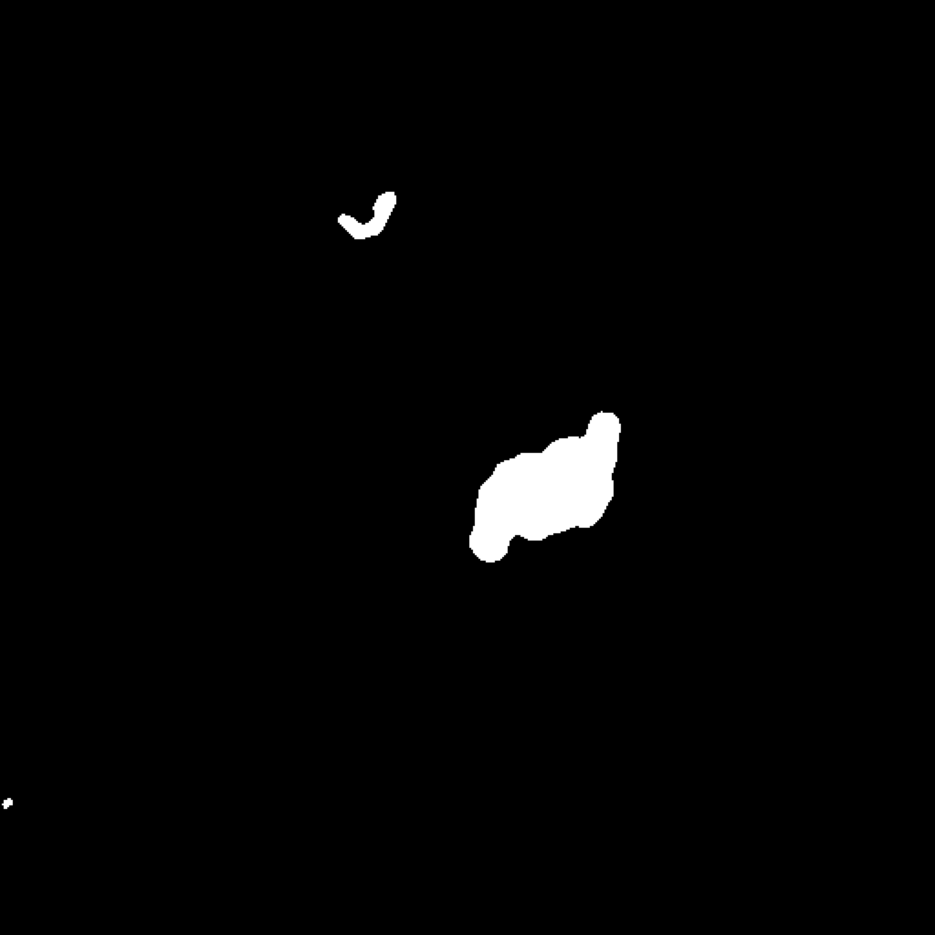}                        
& \includegraphics[width=0.15\linewidth, frame]{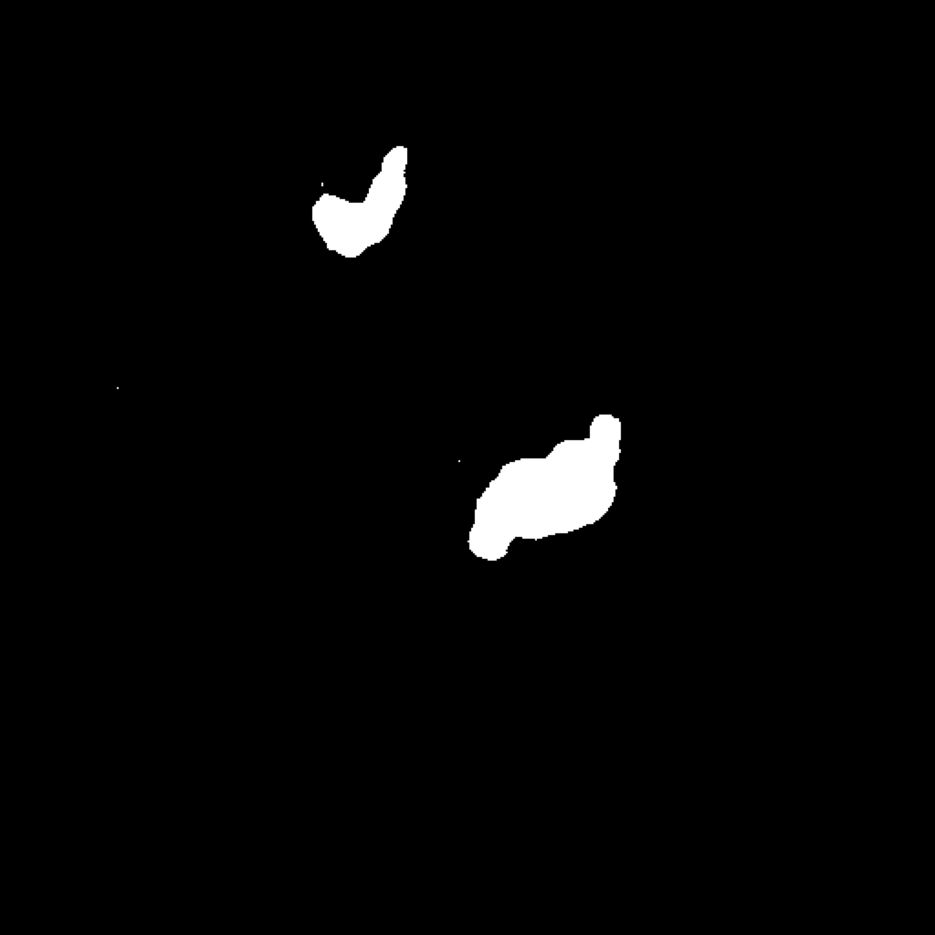}%
\\
\end{tabular}

%% file: tables/examples_segformer.tex
\begin{tabular}{cccccc}
& RGB
& Ground Truth      
& B0
& Magnifier
& B1 
\\[0.2cm]
\rotatebox{90}{\hspace{9mm} MiT}
& \includegraphics[width=0.15\linewidth, frame]{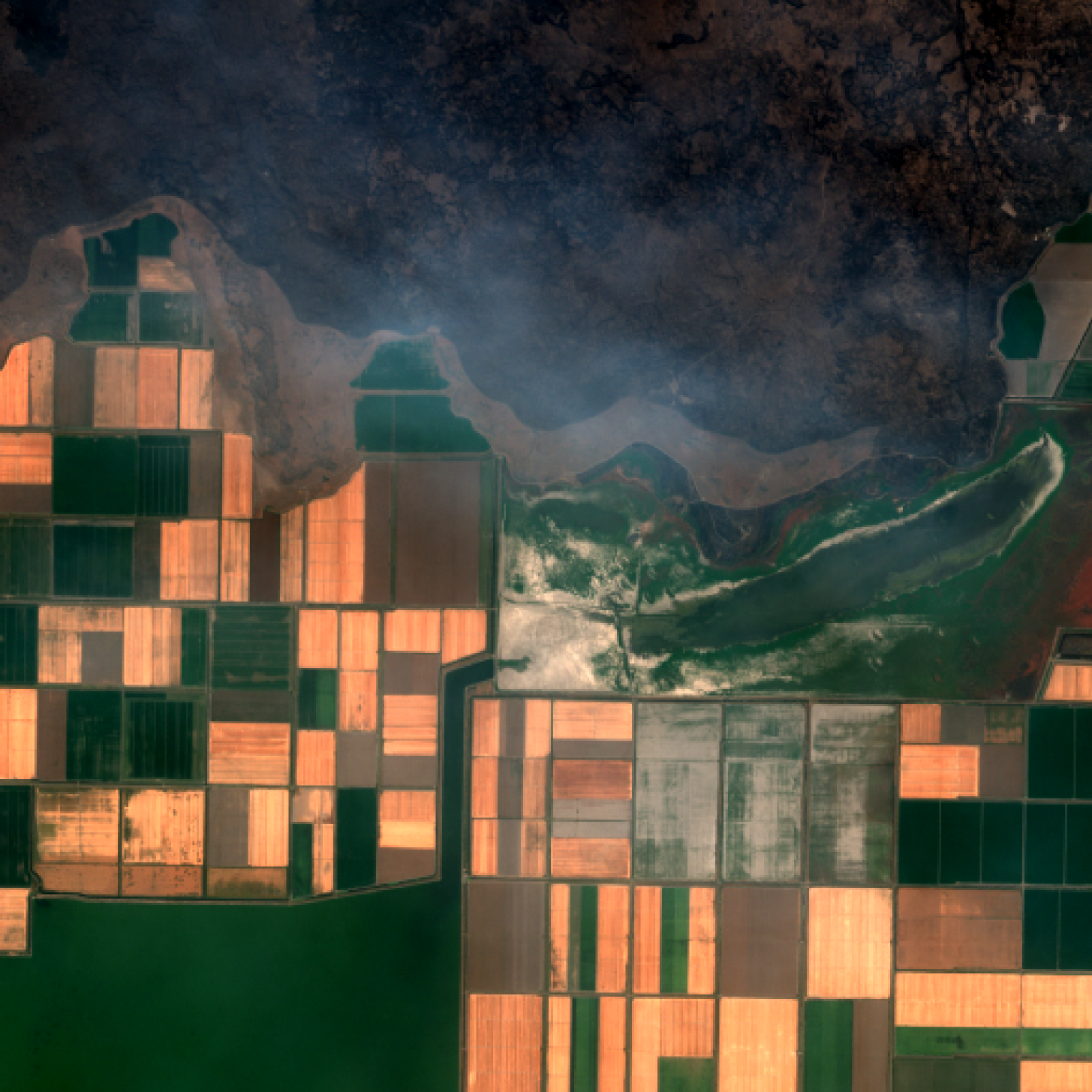}
& \includegraphics[width=0.15\linewidth, frame]{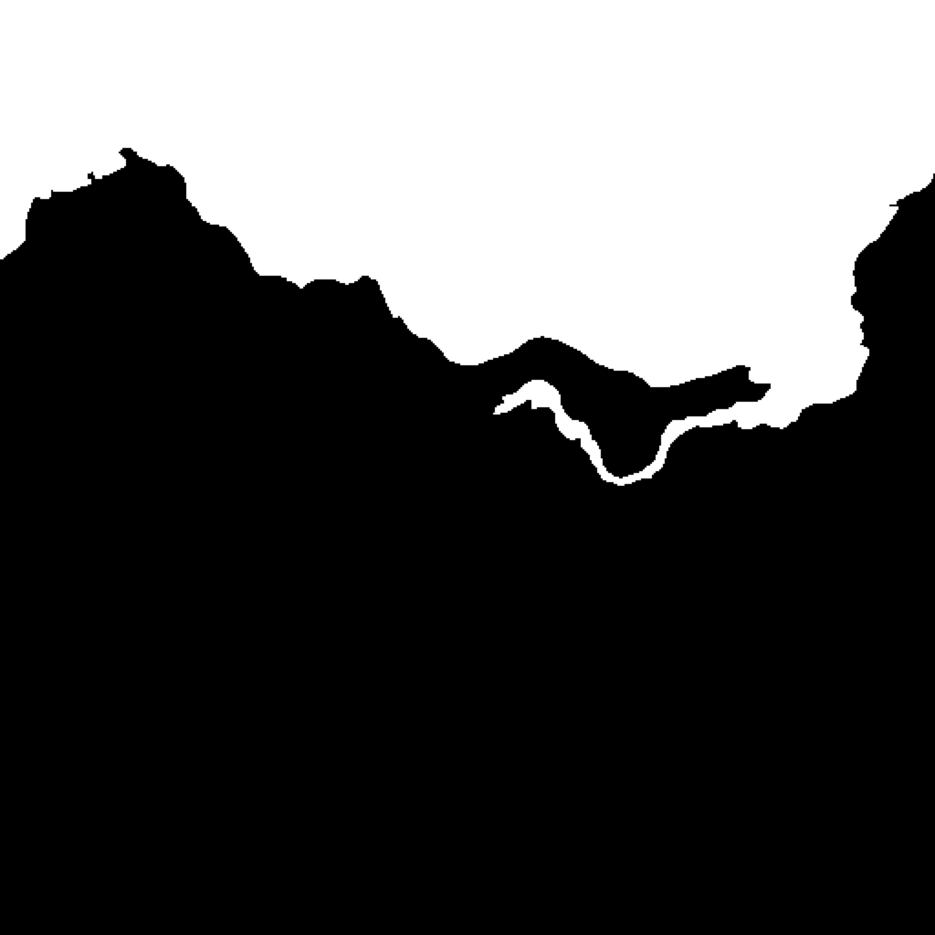}
& \includegraphics[width=0.15\linewidth, frame]{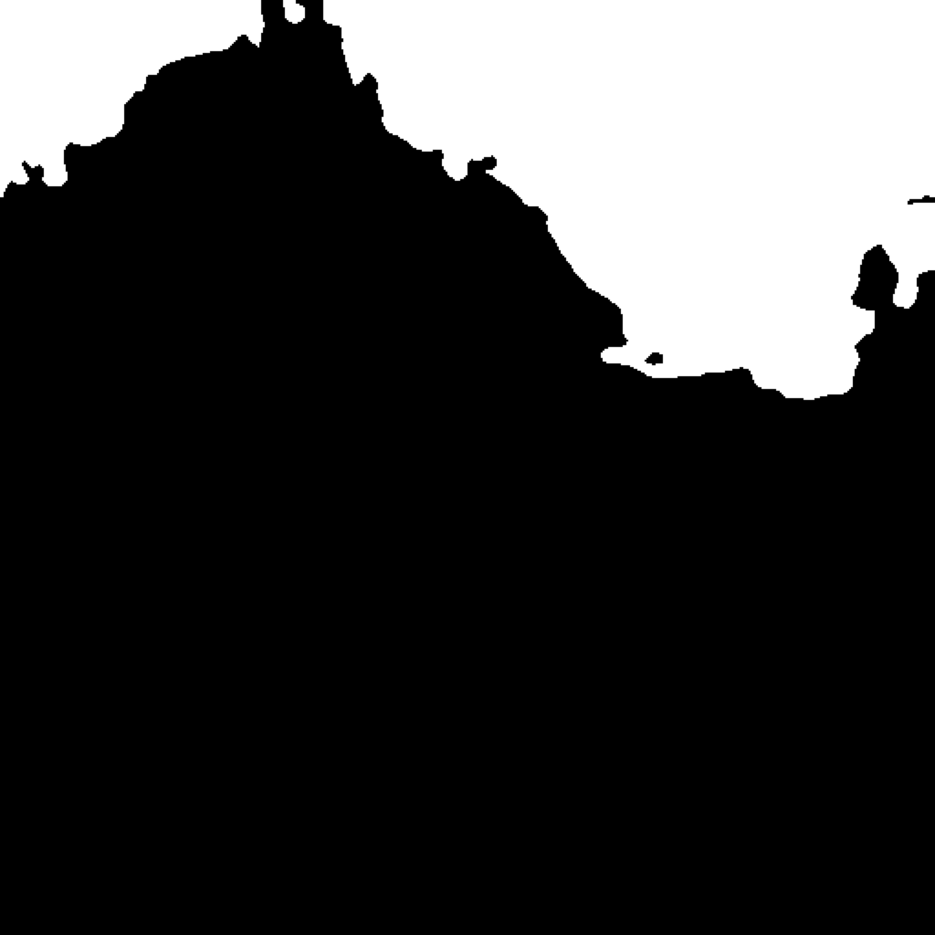}          
& \includegraphics[width=0.15\linewidth, frame]{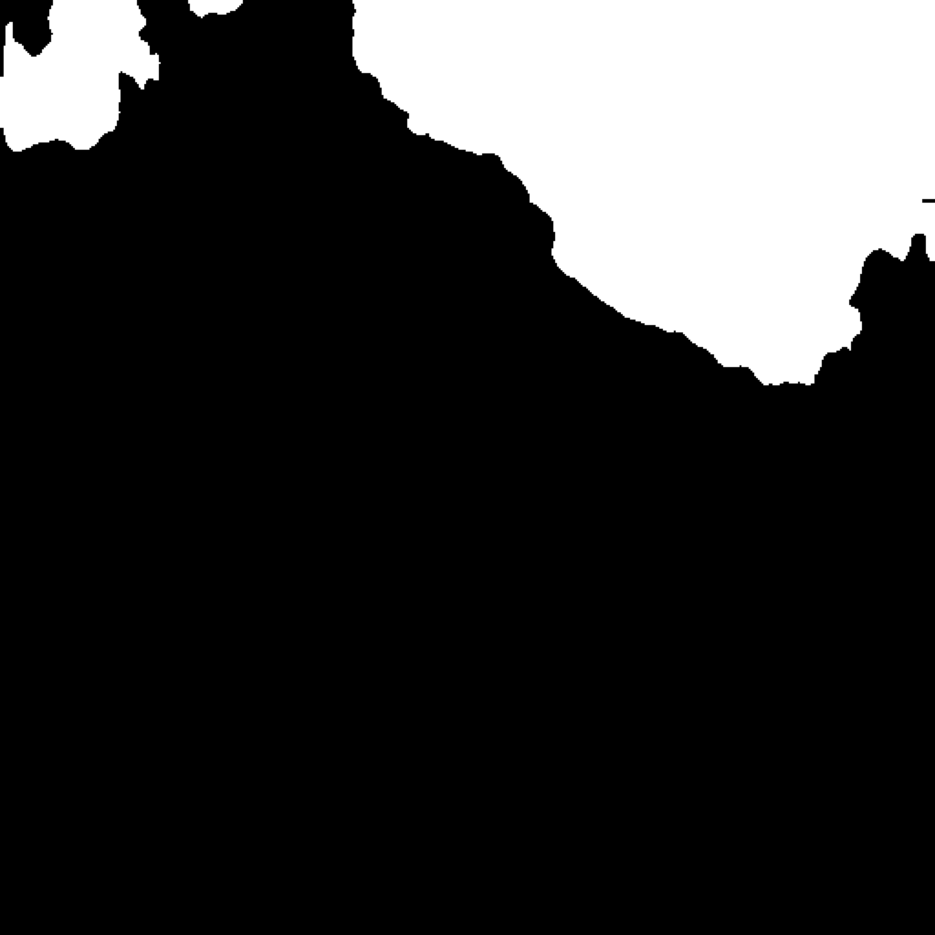}                        
& \includegraphics[width=0.15\linewidth, frame]{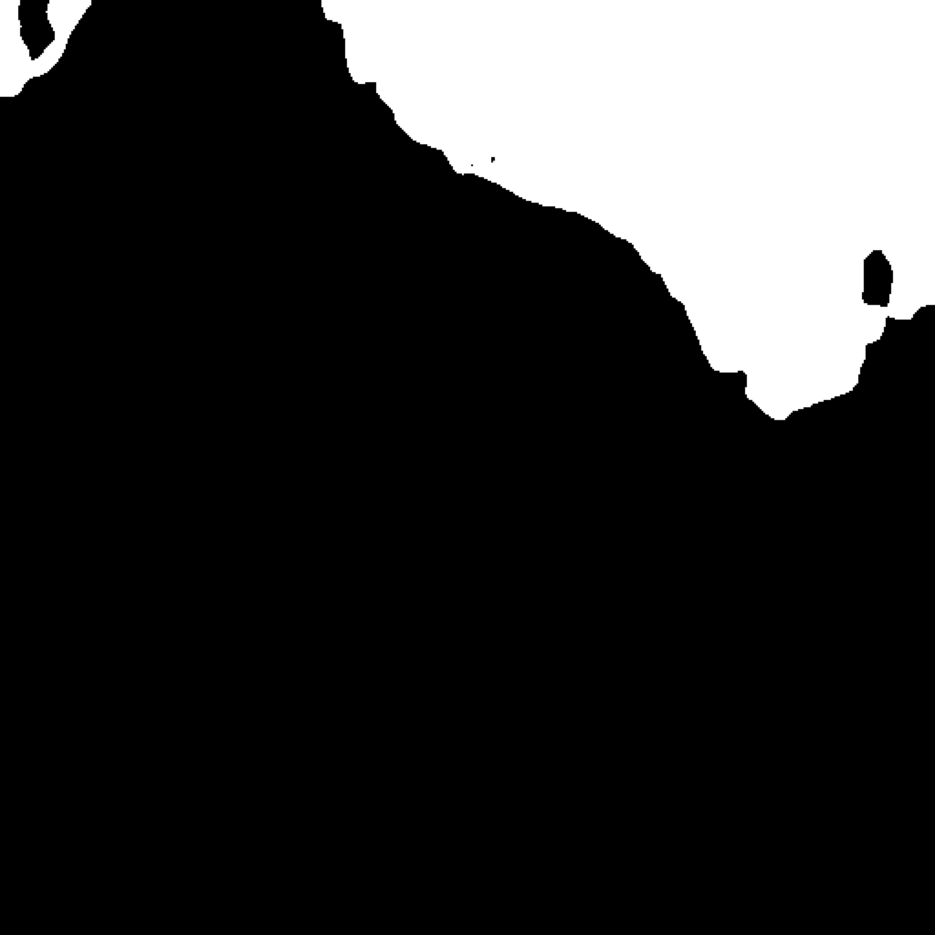}      
\\
\end{tabular}

%% file: paper.bbl
\begin{thebibliography}{10}
\providecommand{\url}[1]{#1}
\csname url@samestyle\endcsname
\providecommand{\newblock}{\relax}
\providecommand{\bibinfo}[2]{#2}
\providecommand{\BIBentrySTDinterwordspacing}{\spaceskip=0pt\relax}
\providecommand{\BIBentryALTinterwordstretchfactor}{4}
\providecommand{\BIBentryALTinterwordspacing}{\spaceskip=\fontdimen2\font plus
\BIBentryALTinterwordstretchfactor\fontdimen3\font minus \fontdimen4\font\relax}
\providecommand{\BIBforeignlanguage}[2]{{%
\expandafter\ifx\csname l@#1\endcsname\relax
\typeout{** WARNING: IEEEtran.bst: No hyphenation pattern has been}%
\typeout{** loaded for the language `#1'. Using the pattern for}%
\typeout{** the default language instead.}%
\else
\language=\csname l@#1\endcsname
\fi
#2}}
\providecommand{\BIBdecl}{\relax}
\BIBdecl

\bibitem{dupuy2020climate}
J.-l. Dupuy, H.~Fargeon, N.~Martin-StPaul, F.~Pimont, J.~Ruffault, M.~Guijarro, C.~Hernando, J.~Madrigal, and P.~Fernandes, ``Climate change impact on future wildfire danger and activity in southern europe: a review,'' \emph{Annals of Forest Science}, vol.~77, no.~2, pp. 1--24, 2020.

\bibitem{ruffault2020increased}
J.~Ruffault, T.~Curt, V.~Moron, R.~M. Trigo, F.~Mouillot, N.~Koutsias, F.~Pimont, N.~Martin-StPaul, R.~Barbero, J.-L. Dupuy \emph{et~al.}, ``Increased likelihood of heat-induced large wildfires in the {M}editerranean {B}asin,'' \emph{Scientific reports}, vol.~10, no.~1, pp. 1--9, 2020.

\bibitem{halofsky2020changing}
\BIBentryALTinterwordspacing
J.~E. Halofsky, D.~L. Peterson, and B.~J. Harvey, ``Changing wildfire, changing forests: the effects of climate change on fire regimes and vegetation in the {P}acific {N}orthwest, {USA},'' \emph{Fire Ecology}, vol.~16, no.~1, pp. 1--26, 2020. [Online]. Available: \url{https://doi.org/10.1186/s42408-019-0062-8}
\BIBentrySTDinterwordspacing

\bibitem{gibson2023image}
R.~K. Gibson, A.~Mitchell, and H.-C. Chang, ``Image texture analysis enhances classification of fire extent and severity using sentinel 1 and 2 satellite imagery,'' \emph{Remote Sensing}, vol.~15, no.~14, p. 3512, 2023.

\bibitem{double_step}
\BIBentryALTinterwordspacing
A.~Farasin, L.~Colomba, and P.~Garza, ``Double-{S}tep {U}-{N}et: {A} {D}eep {L}earning-{B}ased {A}pproach for the {E}stimation of {W}ildfire {D}amage {S}everity through {S}entinel-2 {S}atellite {D}ata,'' \emph{Applied Sciences}, vol.~10, no.~12, 2020. [Online]. Available: \url{https://www.mdpi.com/2076-3417/10/12/4332}
\BIBentrySTDinterwordspacing

\bibitem{california_no_dataset}
D.~Rashkovetsky, F.~Mauracher, M.~Langer, and M.~Schmitt, ``Wildfire detection from multisensor satellite imagery using deep semantic segmentation,'' \emph{IEEE Journal of Selected Topics in Applied Earth Observations and Remote Sensing}, vol.~14, pp. 7001--7016, 2021.

\bibitem{rafik2021}
R.~Ghali, M.~A. Akhloufi, M.~Jmal, W.~Souidene~Mseddi, and R.~Attia, ``Wildfire segmentation using deep vision transformers,'' \emph{Remote Sensing}, vol.~13, no.~17, p. 3527, 2021.

\bibitem{burned_area_delineation}
\BIBentryALTinterwordspacing
L.~Knopp, M.~Wieland, M.~Rättich, and S.~Martinis, ``A {D}eep {L}earning {A}pproach for {B}urned {A}rea {S}egmentation with {S}entinel-2 {D}ata,'' \emph{Remote Sensing}, vol.~12, no.~15, 2020. [Online]. Available: \url{https://www.mdpi.com/2072-4292/12/15/2422}
\BIBentrySTDinterwordspacing

\bibitem{farasin2020supervised}
A.~Farasin, L.~Colomba, G.~Palomba, G.~Nini, and C.~Rossi, ``{S}upervised {B}urned {A}reas delineation by means of {S}entinel-2 imagery and {C}onvolutional {N}eural {N}etworks,'' in \emph{Proceedings of the 17th International Conference on Information Systems for Crisis Response and Management (ISCRAM 2020)}, 2020, pp. 24--27.

\bibitem{palomba2020sentinel}
G.~Palomba, A.~Farasin, and C.~Rossi, ``Sentinel-1 flood delineation with supervised machine learning,'' in \emph{ISCRAM 2020 Conference Proceedings--17th International Conference on Information Systems for Crisis Response and Management}, 2020, pp. 1072--1083.

\bibitem{arnaudo_airquality}
\BIBentryALTinterwordspacing
E.~Arnaudo, A.~Farasin, and C.~Rossi, ``A {C}omparative {A}nalysis for {A}ir {Q}uality {E}stimation from {T}raffic and {M}eteorological {D}ata,'' \emph{Applied Sciences}, vol.~10, no.~13, 2020. [Online]. Available: \url{https://www.mdpi.com/2076-3417/10/13/4587}
\BIBentrySTDinterwordspacing

\bibitem{landcover_dl}
\BIBentryALTinterwordspacing
W.~Zhang, P.~Tang, and L.~Zhao, ``Fast and accurate land-cover classification on medium-resolution remote-sensing images using segmentation models,'' \emph{International Journal of Remote Sensing}, vol.~42, no.~9, pp. 3277--3301, 2021. [Online]. Available: \url{https://doi.org/10.1080/01431161.2020.1871094}
\BIBentrySTDinterwordspacing

\bibitem{landcover_dl2}
\BIBentryALTinterwordspacing
P.~Zhang, Y.~Ke, Z.~Zhang, M.~Wang, P.~Li, and S.~Zhang, ``Urban {L}and {U}se and {L}and {C}over {C}lassification {U}sing {N}ovel {D}eep {L}earning {M}odels {B}ased on {H}igh {S}patial {R}esolution {S}atellite {I}magery,'' \emph{Sensors}, vol.~18, no.~11, 2018. [Online]. Available: \url{https://www.mdpi.com/1424-8220/18/11/3717}
\BIBentrySTDinterwordspacing

\bibitem{zhao2022cnn}
L.~Zhao and S.~Ji, ``Cnn, rnn, or vit? an evaluation of different deep learning architectures for spatio-temporal representation of sentinel time series,'' \emph{IEEE Journal of Selected Topics in Applied Earth Observations and Remote Sensing}, vol.~16, pp. 44--56, 2022.

\bibitem{Datla2021}
\BIBentryALTinterwordspacing
R.~Datla, V.~Chalavadi, and K.~M. C, ``Scene classification in remote sensing images using dynamic kernels,'' in \emph{2021 International Joint Conference on Neural Networks (IJCNN)}.\hskip 1em plus 0.5em minus 0.4em\relax IEEE, Jul. 2021, p. 1–8. [Online]. Available: \url{http://dx.doi.org/10.1109/IJCNN52387.2021.9533648}
\BIBentrySTDinterwordspacing

\bibitem{Datla2024}
\BIBentryALTinterwordspacing
R.~Datla, N.~Perveen, and K.~M. C., ``Learning scene-vectors for remote sensing image scene classification,'' \emph{Neurocomputing}, vol. 587, p. 127679, Jun. 2024. [Online]. Available: \url{http://dx.doi.org/10.1016/j.neucom.2024.127679}
\BIBentrySTDinterwordspacing

\bibitem{unet}
\BIBentryALTinterwordspacing
O.~Ronneberger, P.~Fischer, and T.~Brox, ``U-{N}et: {C}onvolutional {N}etworks for {B}iomedical {I}mage {S}egmentation,'' in \emph{Lecture Notes in Computer Science}.\hskip 1em plus 0.5em minus 0.4em\relax Springer International Publishing, 2015, pp. 234--241. [Online]. Available: \url{https://doi.org/10.1007/978-3-319-24574-4_28}
\BIBentrySTDinterwordspacing

\bibitem{deeplabv3plus}
L.-C. Chen, Y.~Zhu, G.~Papandreou, F.~Schroff, and H.~Adam, ``\BIBforeignlanguage{en}{Encoder-{Decoder} with {Atrous} {Separable} {Convolution} for {Semantic} {Image} {Segmentation}},'' in \emph{\BIBforeignlanguage{en}{Computer {Vision} – {ECCV} 2018}}, ser. Lecture {Notes} in {Computer} {Science}, V.~Ferrari, M.~Hebert, C.~Sminchisescu, and Y.~Weiss, Eds.\hskip 1em plus 0.5em minus 0.4em\relax Cham: Springer International Publishing, 2018, pp. 833--851.

\bibitem{segformer}
\BIBentryALTinterwordspacing
E.~Xie, W.~Wang, Z.~Yu, A.~Anandkumar, J.~M. Alvarez, and P.~Luo, ``Seg{F}ormer: {S}imple and {E}fficient {D}esign for {S}emantic {S}egmentation with {T}ransformers,'' in \emph{Advances in Neural Information Processing Systems}, M.~Ranzato, A.~Beygelzimer, Y.~Dauphin, P.~Liang, and J.~W. Vaughan, Eds., vol.~34.\hskip 1em plus 0.5em minus 0.4em\relax Curran Associates, Inc., 2021, pp. 12\,077--12\,090. [Online]. Available: \url{https://proceedings.neurips.cc/paper/2021/file/64f1f27bf1b4ec22924fd0acb550c235-Paper.pdf}
\BIBentrySTDinterwordspacing

\bibitem{mobilenet}
A.~Howard, M.~Sandler, B.~Chen, W.~Wang, L.-C. Chen, M.~Tan, G.~Chu, V.~Vasudevan, Y.~Zhu, R.~Pang, H.~Adam, and Q.~Le, ``Searching for {MobileNetV3},'' in \emph{2019 {IEEE}/{CVF} {International} {Conference} on {Computer} {Vision} ({ICCV})}, Oct. 2019, pp. 1314--1324, iSSN: 2380-7504.

\bibitem{deep_learning}
\BIBentryALTinterwordspacing
Y.~LeCun, Y.~Bengio, and G.~Hinton, ``Deep learning,'' \emph{Nature}, vol. 521, no. 7553, p. 436–444, May 2015. [Online]. Available: \url{http://dx.doi.org/10.1038/nature14539}
\BIBentrySTDinterwordspacing

\bibitem{cnn}
Y.~Lecun, L.~Bottou, Y.~Bengio, and P.~Haffner, ``Gradient-based learning applied to document recognition,'' \emph{Proceedings of the IEEE}, vol.~86, no.~11, pp. 2278--2324, 1998.

\bibitem{lstm}
\BIBentryALTinterwordspacing
S.~Hochreiter and J.~Schmidhuber, ``Long short-term memory,'' \emph{Neural Comput.}, vol.~9, no.~8, p. 1735–1780, Nov. 1997. [Online]. Available: \url{https://doi.org/10.1162/neco.1997.9.8.1735}
\BIBentrySTDinterwordspacing

\bibitem{brnn}
M.~Schuster and K.~K. Paliwal, ``Bidirectional recurrent neural networks,'' \emph{IEEE transactions on Signal Processing}, vol.~45, no.~11, pp. 2673--2681, 1997.

\bibitem{transformer}
A.~Vaswani, N.~Shazeer, N.~Parmar, J.~Uszkoreit, L.~Jones, A.~N. Gomez, L.~Kaiser, and I.~Polosukhin, ``{A}ttention is {A}ll {Y}ou {N}eed,'' in \emph{Proceedings of the 31st International Conference on Neural Information Processing Systems}, ser. NIPS'17.\hskip 1em plus 0.5em minus 0.4em\relax Red Hook, NY, USA: Curran Associates Inc., 2017, p. 6000–6010.

\bibitem{vit}
\BIBentryALTinterwordspacing
A.~Dosovitskiy, L.~Beyer, A.~Kolesnikov, D.~Weissenborn, X.~Zhai, T.~Unterthiner, M.~Dehghani, M.~Minderer, G.~Heigold, S.~Gelly, J.~Uszkoreit, and N.~Houlsby, ``An {I}mage is {W}orth 16x16 {W}ords: {T}ransformers for {I}mage {R}ecognition at {S}cale,'' 2020. [Online]. Available: \url{https://arxiv.org/abs/2010.11929}
\BIBentrySTDinterwordspacing

\bibitem{bert}
J.~Devlin, M.-W. Chang, K.~Lee, and K.~Toutanova, ``{BERT}: {P}re-training of {D}eep {B}idirectional {T}ransformers for {L}anguage {U}nderstanding,'' in \emph{Proceedings of the 2019 Conference of the North {A}merican Chapter of the Association for Computational Linguistics: Human Language Technologies, Volume 1 (Long and Short Papers)}.\hskip 1em plus 0.5em minus 0.4em\relax Minneapolis, Minnesota: Association for Computational Linguistics, Jun. 2019, pp. 4171--4186.

\bibitem{dqn}
\BIBentryALTinterwordspacing
V.~Mnih, K.~Kavukcuoglu, D.~Silver, A.~A. Rusu, J.~Veness, M.~G. Bellemare, A.~Graves, M.~Riedmiller, A.~K. Fidjeland, G.~Ostrovski, S.~Petersen, C.~Beattie, A.~Sadik, I.~Antonoglou, H.~King, D.~Kumaran, D.~Wierstra, S.~Legg, and D.~Hassabis, ``Human-level control through deep reinforcement learning,'' \emph{Nature}, vol. 518, no. 7540, p. 529–533, Feb. 2015. [Online]. Available: \url{http://dx.doi.org/10.1038/nature14236}
\BIBentrySTDinterwordspacing

\bibitem{otsu1979threshold}
N.~Otsu, ``A threshold selection method from gray-level histograms,'' \emph{IEEE transactions on systems, man, and cybernetics}, vol.~9, no.~1, pp. 62--66, 1979.

\bibitem{normalized_cuts}
J.~Shi and J.~Malik, ``Normalized cuts and image segmentation,'' \emph{IEEE Transactions on Pattern Analysis and Machine Intelligence}, vol.~22, no.~8, pp. 888--905, 2000.

\bibitem{mean_shift}
D.~Comaniciu and P.~Meer, ``Mean shift: a robust approach toward feature space analysis,'' \emph{IEEE Transactions on Pattern Analysis and Machine Intelligence}, vol.~24, no.~5, pp. 603--619, 2002.

\bibitem{tof}
A.~Koudounas, F.~Giobergia, and E.~Baralis, ``Time-of-flight cameras in space: Pose estimation with deep learning methodologies,'' in \emph{2022 IEEE 16th International Conference on Application of Information and Communication Technologies (AICT)}, 2022, pp. 1--6.

\bibitem{deeplab}
L.-C. Chen, G.~Papandreou, I.~Kokkinos, K.~Murphy, and A.~L. Yuille, ``{D}eep{L}ab: {S}emantic {I}mage {S}egmentation with {D}eep {C}onvolutional {N}ets, {A}trous {C}onvolution, and {F}ully {C}onnected {CRF}s,'' \emph{IEEE Transactions on Pattern Analysis and Machine Intelligence}, vol.~40, no.~4, pp. 834--848, 2018.

\bibitem{deeplabv3}
\BIBentryALTinterwordspacing
L.-C. Chen, G.~Papandreou, F.~Schroff, and H.~Adam, ``{R}ethinking {A}trous {C}onvolution for {S}emantic {I}mage {S}egmentation,'' 2017. [Online]. Available: \url{https://arxiv.org/abs/1706.05587}
\BIBentrySTDinterwordspacing

\bibitem{ColombaCikm22}
\BIBentryALTinterwordspacing
L.~Colomba, A.~Farasin, S.~Monaco, S.~Greco, P.~Garza, D.~Apiletti, E.~Baralis, and T.~Cerquitelli, ``A dataset for burned area delineation and severity estimation from satellite imagery,'' in \emph{Proceedings of the 31st ACM International Conference on Information \& Knowledge Management}, ser. CIKM '22.\hskip 1em plus 0.5em minus 0.4em\relax New York, NY, USA: Association for Computing Machinery, 2022, p. 3893–3897. [Online]. Available: \url{https://doi.org/10.1145/3511808.3557528}
\BIBentrySTDinterwordspacing

\bibitem{swin}
Z.~Liu, Y.~Lin, Y.~Cao, H.~Hu, Y.~Wei, Z.~Zhang, S.~Lin, and B.~Guo, ``{S}win {T}ransformer: {H}ierarchical {V}ision {T}ransformer using {S}hifted {W}indows,'' in \emph{Proceedings of the IEEE/CVF International Conference on Computer Vision (ICCV)}, October 2021, pp. 10\,012--10\,022.

\bibitem{resnet}
K.~He, X.~Zhang, S.~Ren, and J.~Sun, ``Deep {Residual} {Learning} for {Image} {Recognition},'' in \emph{2016 {IEEE} {Conference} on {Computer} {Vision} and {Pattern} {Recognition} ({CVPR})}, Jun. 2016, pp. 770--778, iSSN: 1063-6919.

\bibitem{fcn_multiple_res}
S.~Ji, S.~Wei, and M.~Lu, ``Fully convolutional networks for multisource building extraction from an open aerial and satellite imagery data set,'' \emph{IEEE Transactions on Geoscience and Remote Sensing}, vol.~57, no.~1, pp. 574--586, 2019.

\bibitem{bigearthnet}
K.~N. Clasen, L.~Hackel, T.~Burgert, G.~Sumbul, B.~Demir, and V.~Markl, ``reben: Refined bigearthnet dataset for remote sensing image analysis,'' \emph{arXiv preprint arXiv:2407.03653}, 2024.

\bibitem{wang2019scene}
M.~Wang, X.~Zhang, X.~Niu, F.~Wang, and X.~Zhang, ``Scene classification of high-resolution remotely sensed image based on resnet,'' \emph{Journal of Geovisualization and Spatial Analysis}, vol.~3, no.~2, p.~16, 2019.

\bibitem{winter_wheat_mapping}
\BIBentryALTinterwordspacing
J.~Zhang, S.~You, A.~Liu, L.~Xie, C.~Huang, X.~Han, P.~Li, Y.~Wu, and J.~Deng, ``Winter wheat mapping method based on pseudo-labels and u-net model for training sample shortage,'' \emph{Remote Sensing}, vol.~16, no.~14, 2024. [Online]. Available: \url{https://www.mdpi.com/2072-4292/16/14/2553}
\BIBentrySTDinterwordspacing

\bibitem{faqe2023improving}
G.~R. Faqe~Ibrahim, A.~Rasul, and H.~Abdullah, ``Improving crop classification accuracy with integrated sentinel-1 and sentinel-2 data: a case study of barley and wheat,'' \emph{Journal of Geovisualization and Spatial Analysis}, vol.~7, no.~2, p.~22, 2023.

\bibitem{cambrin2024}
\BIBentryALTinterwordspacing
D.~R. Cambrin, E.~Poeta, E.~Pastor, T.~Cerquitelli, E.~Baralis, and P.~Garza, ``Kan you see it? kans and sentinel for effective and explainable crop field segmentation,'' 2024. [Online]. Available: \url{https://arxiv.org/abs/2408.07040}
\BIBentrySTDinterwordspacing

\bibitem{demir2018deepglobe}
I.~Demir, K.~Koperski, D.~Lindenbaum, G.~Pang, J.~Huang, S.~Basu, F.~Hughes, D.~Tuia, and R.~Raskar, ``Deepglobe 2018: A challenge to parse the earth through satellite images,'' in \emph{Proceedings of the IEEE conference on computer vision and pattern recognition workshops}, 2018, pp. 172--181.

\bibitem{Swetha2023}
\BIBentryALTinterwordspacing
G.~Swetha, R.~Datla, C.~Vishnu, and K.~M. C, ``Ms-vacsnet: A network for multi-scale volcanic ash cloud segmentation in remote sensing images,'' in \emph{2023 18th International Conference on Machine Vision and Applications (MVA)}.\hskip 1em plus 0.5em minus 0.4em\relax IEEE, Jul. 2023, p. 1–6. [Online]. Available: \url{http://dx.doi.org/10.23919/MVA57639.2023.10215928}
\BIBentrySTDinterwordspacing

\bibitem{Datla2022}
\BIBentryALTinterwordspacing
R.~Datla, V.~Chalavadi, and K.~M. Chalavadi, ``A multimodal semantic segmentation for airport runway delineation in panchromatic remote sensing images,'' in \emph{Fourteenth International Conference on Machine Vision (ICMV 2021)}, W.~Osten, D.~Nikolaev, and J.~Zhou, Eds.\hskip 1em plus 0.5em minus 0.4em\relax SPIE, Mar. 2022, p.~21. [Online]. Available: \url{http://dx.doi.org/10.1117/12.2622656}
\BIBentrySTDinterwordspacing

\bibitem{vqa_dataset}
X.~Zheng, B.~Wang, X.~Du, and X.~Lu, ``Mutual attention inception network for remote sensing visual question answering,'' \emph{IEEE Transactions on Geoscience and Remote Sensing}, vol.~60, pp. 1--14, 2022.

\bibitem{mmflood}
F.~Montello, E.~Arnaudo, and C.~Rossi, ``Mmflood: A multimodal dataset for flood delineation from satellite imagery,'' \emph{IEEE Access}, vol.~10, pp. 96\,774--96\,787, 2022.

\bibitem{spectral_signature_analysis}
\BIBentryALTinterwordspacing
D.~{van Dijk}, S.~Shoaie, T.~{van Leeuwen}, and S.~Veraverbeke, ``Spectral signature analysis of false positive burned area detection from agricultural harvests using {S}entinel-2 data,'' \emph{International Journal of Applied Earth Observation and Geoinformation}, vol.~97, p. 102296, 2021. [Online]. Available: \url{https://www.sciencedirect.com/science/article/pii/S0303243421000027}
\BIBentrySTDinterwordspacing

\bibitem{stroppiana2012method}
D.~Stroppiana, G.~Bordogna, P.~Carrara, M.~Boschetti, L.~Boschetti, and P.~Brivio, ``A method for extracting burned areas from landsat tm/etm+ images by soft aggregation of multiple spectral indices and a region growing algorithm,'' \emph{ISPRS Journal of Photogrammetry and Remote Sensing}, vol.~69, pp. 88--102, 2012.

\bibitem{martin2006burnt}
M.~P. Mart{\'\i}n, I.~G{\'o}mez, and E.~Chuvieco, ``Burnt area index (baim) for burned area discrimination at regional scale using modis data,'' \emph{Forest Ecology and Management}, no. 234, p. S221, 2006.

\bibitem{bais2}
\BIBentryALTinterwordspacing
F.~Filipponi, ``Bais2: Burned area index for sentinel-2,'' \emph{Proceedings}, vol.~2, no.~7, 2018. [Online]. Available: \url{https://www.mdpi.com/2504-3900/2/7/364}
\BIBentrySTDinterwordspacing

\bibitem{nbr_cit}
D.~Roy, L.~Boschetti, and S.~Trigg, ``Remote sensing of fire severity: assessing the performance of the normalized burn ratio,'' \emph{IEEE Geoscience and Remote Sensing Letters}, vol.~3, no.~1, pp. 112--116, 2006.

\bibitem{nbr2_cit}
\BIBentryALTinterwordspacing
E.~Roteta, A.~Bastarrika, M.~Padilla, T.~Storm, and E.~Chuvieco, ``Development of a {S}entinel-2 burned area algorithm: {G}eneration of a small fire database for sub-{S}aharan {A}frica,'' \emph{Remote Sensing of Environment}, vol. 222, pp. 1--17, 2019. [Online]. Available: \url{https://www.sciencedirect.com/science/article/pii/S0034425718305649}
\BIBentrySTDinterwordspacing

\bibitem{not_unique_thresholds}
\BIBentryALTinterwordspacing
L.~Saulino, A.~Rita, A.~Migliozzi, C.~Maffei, E.~Allevato, A.~P. Garonna, and A.~Saracino, ``{D}etecting {B}urn {S}everity across {M}editerranean {F}orest {T}ypes by {C}oupling {M}edium-{S}patial {R}esolution {S}atellite {I}magery and {F}ield {D}ata,'' \emph{Remote Sensing}, vol.~12, no.~4, 2020. [Online]. Available: \url{https://www.mdpi.com/2072-4292/12/4/741}
\BIBentrySTDinterwordspacing

\bibitem{audebert2017semantic}
N.~Audebert, B.~Le~Saux, and S.~Lef{\`e}vre, ``Semantic {S}egmentation of {E}arth {O}bservation {D}ata {U}sing {M}ultimodal and {M}ulti-scale {D}eep {N}etworks,'' in \emph{Computer Vision -- ACCV 2016}, S.-H. Lai, V.~Lepetit, K.~Nishino, and Y.~Sato, Eds.\hskip 1em plus 0.5em minus 0.4em\relax Cham: Springer International Publishing, 2017, pp. 180--196.

\bibitem{LANG2022112760}
\BIBentryALTinterwordspacing
N.~Lang, N.~Kalischek, J.~Armston, K.~Schindler, R.~Dubayah, and J.~D. Wegner, ``Global canopy height regression and uncertainty estimation from {GEDI} {LIDAR} waveforms with deep ensembles,'' \emph{Remote Sensing of Environment}, vol. 268, p. 112760, 2022. [Online]. Available: \url{https://www.sciencedirect.com/science/article/pii/S0034425721004806}
\BIBentrySTDinterwordspacing

\bibitem{brand2021semantic}
A.~Brand and A.~Manandhar, ``Semantic segmentation of burned areas in satellite images using a {U}-{N}et-based convolutional neural network,'' \emph{The International Archives of Photogrammetry, Remote Sensing and Spatial Information Sciences}, vol.~43, pp. 47--53, 2021.

\bibitem{rs13081509}
\BIBentryALTinterwordspacing
X.~Hu, Y.~Ban, and A.~Nascetti, ``Uni-{T}emporal {M}ultispectral {I}magery for {B}urned {A}rea {M}apping with {D}eep {L}earning,'' \emph{Remote Sensing}, vol.~13, no.~8, 2021. [Online]. Available: \url{https://www.mdpi.com/2072-4292/13/8/1509}
\BIBentrySTDinterwordspacing

\bibitem{HU2023228}
\BIBentryALTinterwordspacing
X.~Hu, P.~Zhang, and Y.~Ban, ``Large-scale burn severity mapping in multispectral imagery using deep semantic segmentation models,'' \emph{ISPRS Journal of Photogrammetry and Remote Sensing}, vol. 196, pp. 228--240, 2023. [Online]. Available: \url{https://www.sciencedirect.com/science/article/pii/S0924271622003410}
\BIBentrySTDinterwordspacing

\bibitem{monaco21}
\BIBentryALTinterwordspacing
S.~Monaco, S.~Greco, A.~Farasin, L.~Colomba, D.~Apiletti, P.~Garza, T.~Cerquitelli, and E.~Baralis, ``{A}ttention to {F}ires: {M}ulti-{C}hannel {D}eep {L}earning {M}odels for {W}ildfire {S}everity {P}rediction,'' \emph{Applied Sciences}, vol.~11, no.~22, 2021. [Online]. Available: \url{https://www.mdpi.com/2076-3417/11/22/11060}
\BIBentrySTDinterwordspacing

\bibitem{burntnet}
\BIBentryALTinterwordspacing
S.~T. Seydi, M.~Hasanlou, and J.~Chanussot, ``Burnt-net: Wildfire burned area mapping with single post-fire sentinel-2 data and deep learning morphological neural network,'' \emph{Ecological Indicators}, vol. 140, p. 108999, 2022. [Online]. Available: \url{https://www.sciencedirect.com/science/article/pii/S1470160X22004708}
\BIBentrySTDinterwordspacing

\bibitem{cabuar}
D.~Rege~Cambrin, L.~Colomba, and P.~Garza, ``Ca{B}u{A}r: California {B}urned {A}reas dataset for delineation,'' \emph{IEEE Geoscience and Remote Sensing Magazine}, 2023.

\bibitem{prabowo2022}
Y.~Prabowo, A.~D. Sakti, K.~A. Pradono, Q.~Amriyah, F.~H. Rasyidy, I.~Bengkulah, K.~Ulfa, D.~S. Candra, M.~T. Imdad, and S.~Ali, ``Deep learning dataset for estimating burned areas: Case study, indonesia,'' \emph{Data}, vol.~7, no.~6, p.~78, 2022.

\bibitem{sentinel2}
\BIBentryALTinterwordspacing
M.~Drusch, U.~D. Bello, S.~Carlier, O.~Colin, V.~Fernandez, F.~Gascon, B.~Hoersch, C.~Isola, P.~Laberinti, P.~Martimort, A.~Meygret, F.~Spoto, O.~Sy, F.~Marchese, and P.~Bargellini, ``Sentinel-2: {ESA}{\textquotesingle}s optical high-resolution mission for {GMES} operational services,'' \emph{Remote Sensing of Environment}, vol. 120, pp. 25--36, May 2012. [Online]. Available: \url{https://doi.org/10.1016/j.rse.2011.11.026}
\BIBentrySTDinterwordspacing

\bibitem{l2a}
\BIBentryALTinterwordspacing
``Sentinel-2 l2a user guide,'' 2023. [Online]. Available: \url{https://sentinel.esa.int/web/sentinel/user-guides/sentinel-2-msi/product-types/level-2a}
\BIBentrySTDinterwordspacing

\bibitem{roy2014}
D.~P. Roy, M.~A. Wulder, T.~R. Loveland, C.~E. Woodcock, R.~G. Allen, M.~C. Anderson, D.~Helder, J.~R. Irons, D.~M. Johnson, R.~Kennedy \emph{et~al.}, ``Landsat-8: Science and product vision for terrestrial global change research,'' \emph{Remote sensing of Environment}, vol. 145, pp. 154--172, 2014.

\bibitem{yeung_2022}
\BIBentryALTinterwordspacing
M.~Yeung, E.~Sala, C.-B. Schönlieb, and L.~Rundo, ``Unified focal loss: {G}eneralising dice and cross entropy-based losses to handle class imbalanced medical image segmentation,'' \emph{Computerized Medical Imaging and Graphics}, vol.~95, p. 102026, 2022. [Online]. Available: \url{https://www.sciencedirect.com/science/article/pii/S0895611121001750}
\BIBentrySTDinterwordspacing

\bibitem{janez2006}
J.~Dem\v{s}ar, ``Statistical comparisons of classifiers over multiple data sets,'' \emph{J. Mach. Learn. Res.}, vol.~7, p. 1–30, Dec. 2006.

\bibitem{cross-validation}
\BIBentryALTinterwordspacing
P.~Refaeilzadeh, L.~Tang, and H.~Liu, \emph{Cross-Validation}.\hskip 1em plus 0.5em minus 0.4em\relax Springer US, 2009, p. 532–538. [Online]. Available: \url{http://dx.doi.org/10.1007/978-0-387-39940-9_565}
\BIBentrySTDinterwordspacing

\end{thebibliography}
